\documentclass[journal]{IEEEtran}
\usepackage{balance}
\usepackage{amsmath}
\usepackage{array}
\usepackage{graphicx}
\usepackage{amssymb}
\usepackage{pifont}
\usepackage{subcaption}
\usepackage{booktabs}
\usepackage{multirow}
\usepackage{color}
\usepackage{xcolor}
\usepackage{cite}
\usepackage{bm}
\usepackage{framed}
\newcommand{\cmark}{\ding{51}}
\newcommand{\xmark}{\ding{55}}
\usepackage{makecell}
\usepackage{colortbl}
\usepackage{algorithm}
\usepackage{algorithmic}
\usepackage{balance}

\ifCLASSINFOpdf

\else

\fi

\begin{document}
\title{FAMINet: Learning Real-time Semi-supervised \\ Video Object Segmentation with \\ Steepest Optimized Optical Flow}

\author{Ziyang Liu,
		Jingmeng Liu,
		Weihai Chen*,
		Xingming Wu, and
        Zhengguo Li
\thanks{*Corresponding author: Weihai Chen, whchen@buaa.edu.cn.}
\thanks{Z. Liu, J. Liu, W. Chen and X. Wu are with School of Automation Science and Electrical Engineering, Beihang University, 100191, Beijing, China.}
\thanks{Z. Li is with SRO Department, Institute for Infocomm Research, Singapore, 138632, Singapore.}
}

\maketitle

\begin{abstract}
\textcolor{black}{Semi-supervised video object segmentation (VOS) aims to segment a few moving objects in a video sequence, where these objects are specified by annotation of first frame. The optical flow has been considered in many existing semi-supervised VOS methods to improve the segmentation accuracy. However, the optical flow-based semi-supervised VOS methods cannot run in real time due to high complexity of optical flow estimation. A FAMINet, which consists of a feature extraction network (F), an appearance network (A), a motion network (M), and an integration network (I), is proposed in this study to address the abovementioned problem. The appearance network outputs an initial segmentation result based on static appearances of objects. The motion network estimates the optical flow via very few parameters, which are optimized rapidly by an online memorizing algorithm named relaxed steepest descent. The integration network refines the initial segmentation result using the optical flow. Extensive experiments demonstrate that the FAMINet outperforms other state-of-the-art semi-supervised VOS methods on the DAVIS and YouTube-VOS benchmarks, and it achieves a good trade-off between accuracy and efficiency. Our code is available at \textit{https://github.com/liuziyang123/FAMINet}.}
\end{abstract}


\IEEEpeerreviewmaketitle

\section{Introduction}
\textcolor{black}{Video object segmentation (VOS) is supposed to segment a few moving objects in a separate video sequence. The VOS is crucial for various image and video processing tasks \cite{perazzi2016benchmark}, such as interactive video editing, surveillance, and augmented reality, and for many robotic tasks, including autonomous driving \cite{pei2021multifeature}, obstacle avoidance, and 3D semantic mapping \cite{qiu2018rgb}.} The VOS has always been \textcolor{black}{a major research area} in the \textcolor{black}{field} of computer vision due to its application value. \textcolor{black}{The VOS has three types}, including unsupervised, semi-supervised, and interactive. This \textcolor{black}{study} focuses on the most challenging semi-supervised VOS, where the objects that need to be segmented in a video are specified by the \textcolor{black}{annotation of the first frame} \cite{perazzi2016benchmark}. The VOS refers to the semi-supervised VOS in the following text unless otherwise stated.

\begin{figure}[t]
	\centering
	\includegraphics[width=80mm]{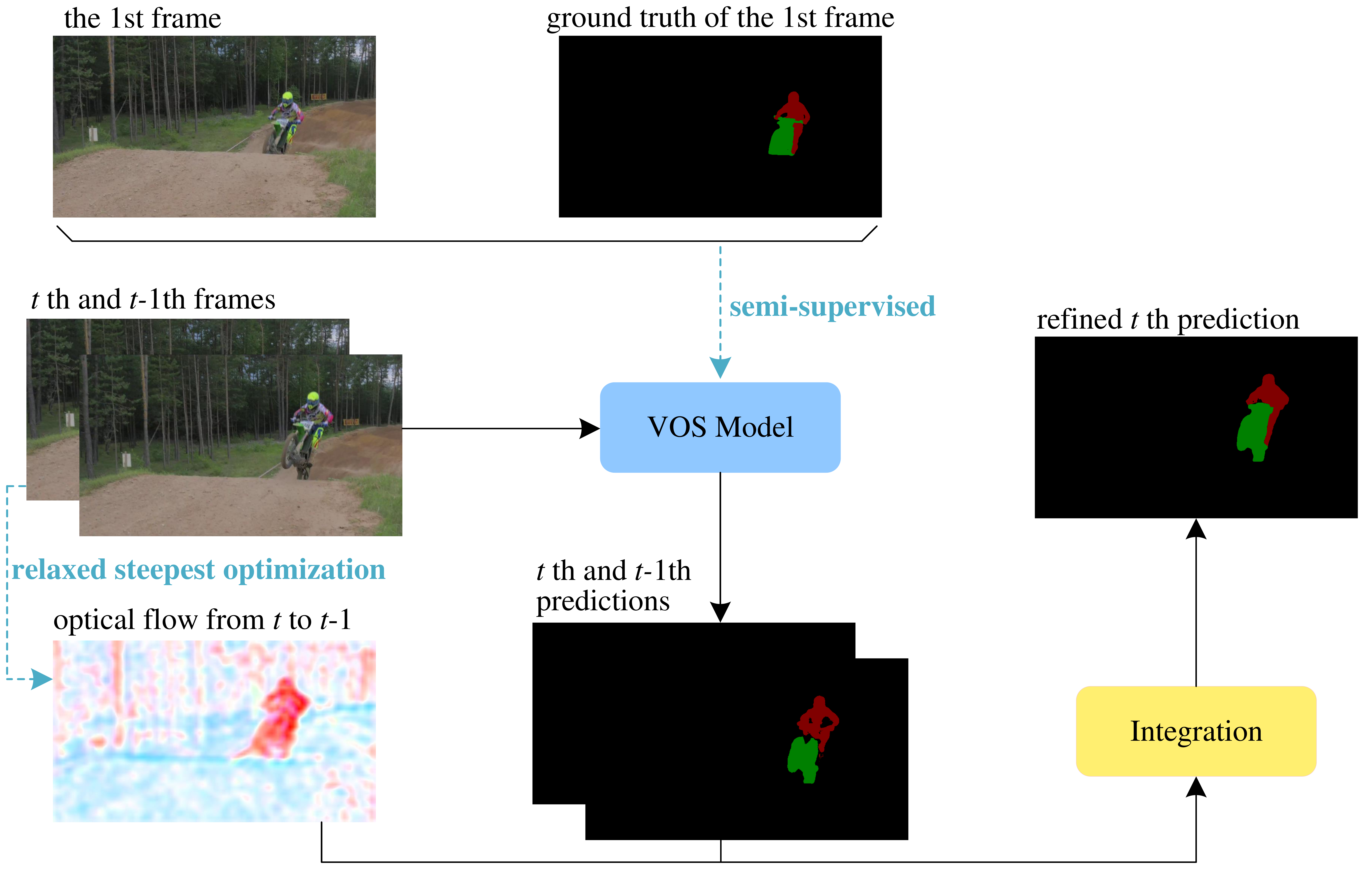}
	\caption{\textcolor{black}{Overall pipeline of the FAMINet. Given the first frame and its ground truth annotation, the semi-supervised VOS model learns to segment target objects in consecutive video frames. We propose to adopt the optical flow, which is optimized by the RSD online, to refine the current prediction by incorporating temporal information of previous frames. Such an optical flow is sufficiently accurate in the foreground object part.}}
	\label{fig:pipeline}
\end{figure}

Remarkable progress has been made in the VOS with the development of deep learning, where a convolutional neural network (CNN) is usually fine-tuned on the first annotated frame for segmenting target objects \cite{caelles2017one, robinson2020learning}. \textcolor{black}{Notably,} most target objects are in rigid or nonrigid motions relative to the camera throughout videos, which brings great challenges to the deep learning-based VOS methods. The CNN may fail in segmenting the objects when their placements, shapes, and appearances in subsequent frames are very different from those in the first frame. Fortunately, the \textcolor{black}{motions of the objects} can be represented by optical flow, which records pixel-wise displacements between consecutive frames \cite{teed2020raft}. The optical flow provides a prior knowledge about the moving objects and \textcolor{black}{can thus} be an important cue for assisting the VOS.

\textcolor{black}{Many studies have incorporated the optical flow into the VOS.} Some of these methods input the optical flow or its high-level feature representations into VOS networks to improve their performances \cite{cheng2017segflow, jain2017fusionseg, tokmakov2017learning, perazzi2017learning, khoreva2019lucid}. Other methods directly convert the optical flow into the final segmentation result via the CNN end-to-end \cite{tokmakov2017learning2, hu2020motion}. \textcolor{black}{In the VOS, the optical flow performs well against the problems caused by movements and deformations of objects (e.g., the problems of occlusion, changes in object shapes, and changes in object appearances).}

In most of these optical flow-based VOS methods, the optical flow is assumed to be available. \textcolor{black}{However}, obtaining the optical flow \textcolor{black}{in practice} is often a time-consuming task, which has to be completed by complicated traditional methods \cite{brox2010large, revaud2015epicflow} or by deep CNNs \cite{dosovitskiy2015flownet, ilg2017flownet, teed2020raft}.
Therefore, introducing the optical flow into the VOS can bring lots of extra computations, which \textcolor{black}{cannot} be applied to real-time systems.

\textcolor{black}{To address this problem, a FAMINet is proposed for the VOS with the help of optical flow via a fast inference speed in this paper.} The FAMINet is composed of a feature extraction network (F), an appearance network (A), a motion network (M), and an integration network (I). The feature extraction network \textcolor{black}{extracts} \textcolor{black}{multi level} feature representations from video frames and supply them to other network components. The appearance network outputs an initial segmentation result based on \textcolor{black}{static appearances of objects}. It can be implemented by existing state-of-the-art VOS methods, \textcolor{black}{such as} the FRTM-fast in \cite{robinson2020learning}. {\it{The motion network is a distinctive component and its parameters are optimized online by a novel relaxed steepest descent (RSD) algorithm proposed in this \textcolor{black}{study}.}} It is a lightweight network that can estimate the optical flow efficiently. For online optimization, the objective function is determined according to the unsupervised learning scheme in \cite{jonschkowski2020matters}. The optimization algorithm adopts the proposed RSD, which leads to a significantly faster convergence speed than commonly used stochastic gradient descent (SGD) algorithm. \textcolor{black}{The motion network can produce sufficiently accurate optical flow with high efficiency owing to the very few network parameters along with the fast RSD optimization.} {\it{The integration network is another distinctive component proposed in this paper.}} It takes the initial segmentation from appearance network, optical flow from motion network, and previous frames' predictions as inputs, and \textcolor{black}{it} outputs the final refined segmentation result. The integration network is also lightweight. The overall pipeline of FAMINet is shown in Figure \ref{fig:pipeline}.

Compared with \cite{caelles2017one, robinson2020learning}, which segment video frames independently, the FAMINet exploits inter-frame temporal information by introducing the optical flow to facilitate more temporal-consistent segmentation results. Thus, it is more powerful in dealing with small objects, occlusions, \textcolor{black}{and object changes}. The FAMINet is also different from existing optical flow-based VOS methods \cite{cheng2017segflow, jain2017fusionseg, tokmakov2017learning, perazzi2017learning, khoreva2019lucid}. Its motion network estimates the optical flow in a much more time-saving way. Its integration network incorporates multi-frame's predictions according to the pixel-wise correspondence provided by optical flow, which is more explainable than simply inputting the optical flow or its features into VOS networks. Overall, \textcolor{black}{the main contributions of this study are as follows}:
\begin{itemize}
\item A novel CNN-based method named FAMINet is proposed to learn the VOS with optical flow in this \textcolor{black}{study}, which achieves a good trade-off between accuracy and efficiency compared \textcolor{black}{with} state-of-the-art VOS methods.
\item A lightweight motion network along with an RSD online optimization algorithm are proposed to estimate the optical flow efficiently. The RSD algorithm can also be applied to optimize other objective functions with a fast convergence speed.
\item \textcolor{black}{An integration network is proposed to achieve robust segmentations of objects despite of the problems caused by movements and deformations of objects, given the optical flow estimated by the motion network.}

\end{itemize}

\section{Related work}
In this section, existing VOS methods with and without optical flow are reviewed.

\subsection{\textcolor{black}{Optical flow-free VOS}}
Recently, most VOS methods are deep learning-based. \textcolor{black}{Given the annotation of the first frame, the CNN has different ways in determining which target objects to be segmented.}
\subsubsection{Online fine-tuning}
The CNN is trained on large-scale datasets offline, \textcolor{black}{and then, it is} fine-tuned on the first frame for segmenting the target objects online \cite{caelles2017one, xu2018youtube, maninis2018video}. \textcolor{black}{However,} fine-tuning on only one frame can lead to \textcolor{black}{overfitting of the CNN} and degrade \textcolor{black}{generic segmentation ability of the CNN} learned offline. This problem can be eased by fine-tuning on subsequent video frames \cite{voigtlaender2017online}, fine-tuning only a small part of the CNN \cite{robinson2020learning}\textcolor{black}{,} or data augmentation \cite{khoreva2019lucid}.
\subsubsection{Mask propagation}
The target objects are specified by \textcolor{black}{predicted segmentation mask of previous frame} \cite{perazzi2017learning}. This method is well adapted to \textcolor{black}{continuous changes in objects}, but \textcolor{black}{it} can encounter troubles at occlusions. Error accumulation is another serious problem \textcolor{black}{given that prediction of current frame} depends on previous ones. Several methods work on incorporating the initial ground-truth mask to avoid these problems \cite{oh2018fast, yang2018efficient, wang2019ranet, johnander2019generative}. Incorporating previous multi frames instead of a single one is also a good solution \cite{oh2019video}.
\subsubsection{Feature matching}
Feature matching is a bit complicated way to specify the target objects \cite{voigtlaender2019feelvos}. Features extracted from the current frame are matched with those from the \textcolor{black}{masked areas of the first frame}. Then\textcolor{black}{,} the objects can be correctly determined according to the feature correspondence. Such a paradigm makes the CNN trainable end-to-end. 

{\color{black}Different from optical flow-free methods, the proposed FAMINet incorporates optical flow to extract temporal information for the VOS against the problems of occlusions, changes in object shapes , and changes in object appearances. The FAMINet segments objects dynamically (where video frames are segmented jointly by optical flow) rather than statically (where each video frame is segmented independently).
}

\begin{figure*}[t]
	\centering
	\includegraphics[width=160mm]{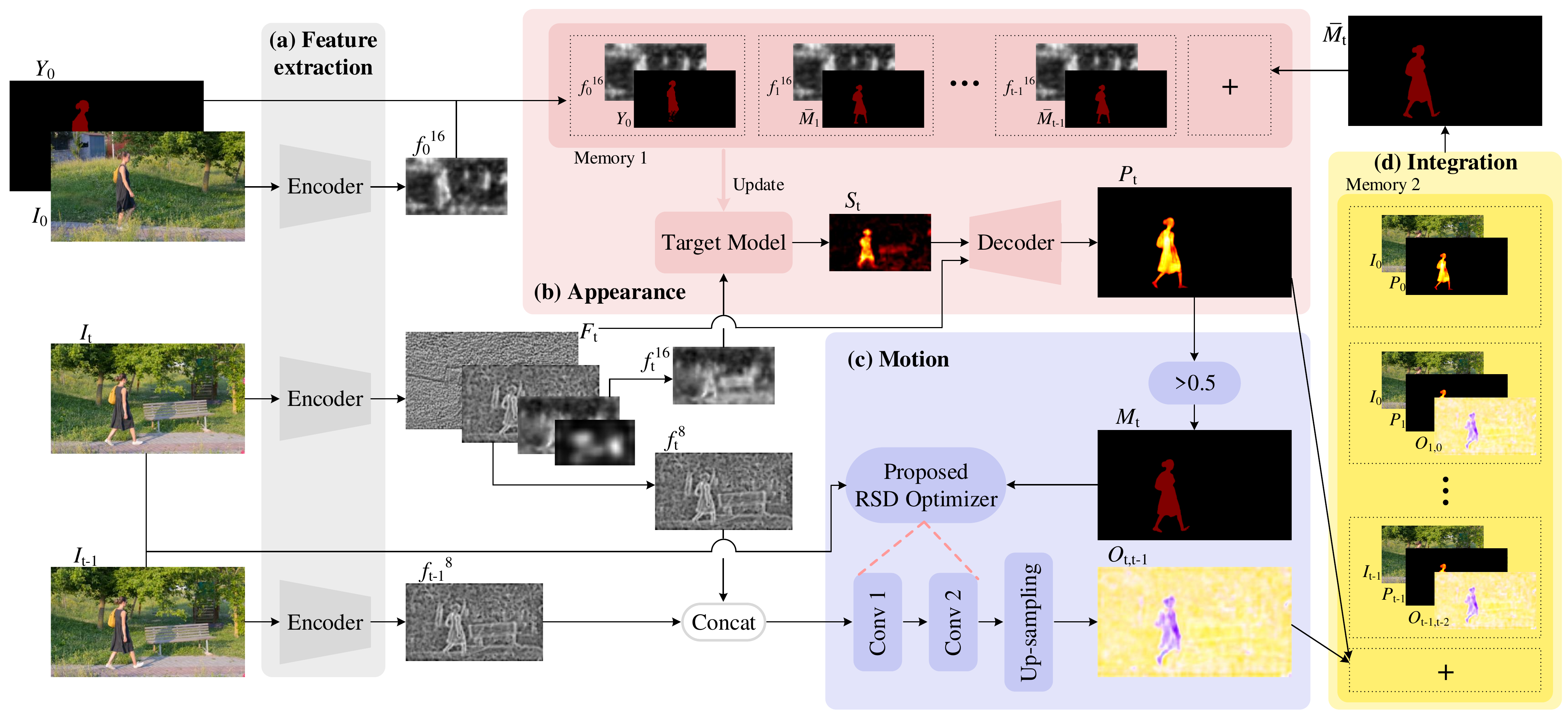}
	\caption{The FAMINet consists of a feature extraction network, an appearance network, a motion network, and an integration network. The parameters of encoders in the feature extraction network are {\it{shared}}. Memory1 and Memory2 indicate the CPU memories, which store variables temporarily.}
	\label{fig:framework}
\end{figure*}

\subsection{Optical flow-based VOS}
\subsubsection{Inputting optical flow to VOS}
A few optical flow-based VOS methods extract high-level feature representations from the optical flow and merge them into VOS networks, to implicitly learn the motion information \cite{cheng2017segflow, jain2017fusionseg, khoreva2019lucid}. A more explicit learning method is to incorporate the optical flow with {\color{black}the feature representations of current and previous frames} via a convolutional gated recurrent unit (ConvGRU), by which segmentation results are smoothed across frames and maintain temporally consistent \cite{tokmakov2017learning}. The optical flow can also be utilized to propagate previous object masks for refining the current segmentation result \cite{luiten2018premvos}.
\subsubsection{Converting optical flow to VOS}
Besides providing {\color{black}motion information of objects}, {\color{black}the optical flow depicts the rough contours of moving objects \cite{perazzi2017learning}.} Thus, the magnitude of optical flow is computed and fed into the VOS network to improve its performance.  The same idea is further extended by \cite{tokmakov2017learning2, hu2020motion}, where the optical flow is directly converted into the final object mask based on its description of {\color{black}object shapes}.
\subsubsection{Other methods}
In \cite{tsai2016video}, the VOS and optical flow are iteratively optimized, where the VOS is realized by a optical flow-based graphical model. In \cite{xie2020moving}, the moving objects are divided into active and passive moving ones according to the optical flow. Then, the passive moving objects can be segmented by a simple calculation.

{\color{black}Optical flow-based methods can achieve robust object segmentation, but they are very time-consuming due to high complexity of optical flow estimation. Different from these methods, the optical flow is estimated through a very fast speed in the motion network of FAMINet. The optical flow is optimized by the proposed RSD algorithm online via only few iterations. With the help of the fast optimized optical flow, the FAMINet achieves robust object segmentation while significantly surpassing optical flow-based methods in terms of running speed.
}

\section{Method}
In this section, the FAMINet is proposed for learning the VOS with optical flow. It includes a feature extraction network, an appearance network, a motion network, and an integration network, as shown in Figure \ref{fig:framework}. {\color{black}Notably,} the feature extraction and appearance networks are implemented by existing technologies, while {\it{the motion and integration networks are two distinctive components proposed in this {\color{black}study}.}}

\subsection{Feature extraction network}
Input video frames are {\color{black}first} mapped into feature domain by {\color{black}an} encoder $\mathcal{F}_{\theta}$. The $t$th frame $I_t$ is encoded into feature $f_t^s$ through mapping $\mathcal{F}_{\theta}: \mathbb{R}^{h\times{w}\times{3}} \rightarrow  \mathbb{R}^{{\frac{h}{s}}\times{\frac{w}{s}}\times{c}}$, where $h$ and $w$ are the height and width of $I_t$. {\color{black}Notably,} $c$ and $s$ correspond to the channel dimension of $f_t^s$ and the scale of $f_t^s$ relative to $I_t$, respectively. Multi-scale features $F_t$ is {\color{black}composed} of $f_t^s$ at different scales (e.g., $s=4, 8, 16, 32$). The extracted features are then supplied to the appearance and motion networks.

The encoder $\mathcal{F}_{\theta}$ is implemented by ResNet \cite{he2016deep}. Its parameters are pre-trained on ImageNet \cite{russakovsky2015imagenet} and then kept {\it{fixed}} during training and evaluating the FAMINet.

\subsection{Appearance network}
The appearance network outputs an initial segmentation result based on {\color{black}static appearances of objects}. It is implemented by a state-of-the-art VOS method, the FRTM-fast \cite{robinson2020learning}, which is composed of a target model $\mathcal{T}_{\tau}$ and a decoder network $\mathcal{D}_{\theta}$. For the $t$th frame $I_t$, the target object to be segmented is discriminated by $\mathcal{T}_{\tau}$, which takes in frame's single-scale feature $f_t^{16}$ and produces an object-aware score map $S_t$. The coarse score map $S_t$ is then refined into high-resolution object probability map $P_t$ by $\mathcal{D}_{\theta}$, which takes in frame's multi-scale features $F_t$. The computations can be represented by:
\begin{equation}
P_t = \mathcal{D}_{\theta}(F_t, \mathcal{T}_{\tau}(f_t^{16})),
\label{eq:appearancecomputation}
\end{equation}
where the parameters $\tau$ are optimized by Gauss-Newton solvers online, given the first frame's single-scale feature $f_0^{16}$ and {\color{black}ground truth} object annotation $Y_0$.
$\mathcal{D}_{\theta}$ is trained on large-scale datasets offline.

{\color{black}Notably,} the appearance network can be implemented by other VOS methods. We choose the FRTM-fast
as it can achieve high-accuracy segmentation results with a fast inference speed. However, the FRTM-fast segments videos frame by frame without utilizing temporal information inter frames. {\color{black}In the next part,} we propose the motion and integration networks to incorporate the temporal information, which further improves the segmentation accuracy via few additional computations.

\subsection{Motion network}
\label{sec:motion}
The motion network {\color{black}estimates} the optical flow efficiently. Let $O_{t,t-1}$ {\color{black}denote} the optical flow from frame $I_t$ to $I_{t-1}$. Estimating $O_{t,t-1}$ is often computationally {\color{black}costly} \cite{ilg2017flownet, teed2020raft, ranjan2017optical, sun2018pwc}.
Introducing the optical flow into the VOS can greatly slow the inference speed, which {\color{black}does} not achieve a good trade-off between accuracy and efficiency \cite{cheng2017segflow, jain2017fusionseg, tokmakov2017learning, perazzi2017learning}. Therefore, we expect that the motion network can produce VOS-oriented optical flow via as few as possible computations.

{\color{black}For efficiency}, only two convolutional layers are utilized to estimate the optical flow in the motion network:
\begin{equation}
O_{t, t-1} = \mathcal{O}_{\omega}(f_{t, t-1}^s) = \uparrow (\omega^{(2)} * (\omega^{(1)} * f_{t, t-1}^s)),
\label{eq:opticalflow}
\end{equation}
where $f_{t, t-1}^s$ is the concatenation of single-scale features, $f_t^s$ and $f_{t-1}^s$, from two consecutive frames. $*$ and $\uparrow$ denote convolutional operation and up-sampling, respectively. $\mathcal{O}_{\omega}$ is a lightweight linear model, which takes up few computations.

{\color{black}Notably,} {\color{black}ground truth} label for dense optical flow is absent in non-optical flow datasets (e.g., the VOS datasets). Thus, to train $\mathcal{O}_{\omega}$, an unsupervised learning scheme is introduced {\color{black}on the basis of} the assumption of photometric consistency \cite{jonschkowski2020matters}.
Given a non-parametric warping function based on $\textit{grid sample}$ \cite{jason2016back}, $I_t$ can be reconstructed from $I_{t-1}$ by $O_{t, t-1}$:
\begin{equation}
\tilde{I_t}(u,v) = \mathcal{W}(I_{t-1}(u,v), O_{t, t-1}(u,v)),
\label{eq:warp}
\end{equation}
where $(u,v)$ represents the pixel coordinate. $\mathcal{W}$ is the warping function and $\tilde{I_t}$ is the reconstructed image. {\color{black}Notably,} the warping direction is opposite to the flow direction. The objective function is formalized as the dissimilarity between the reconstructed frame $\tilde{I_t}$ and original frame $I_{t}$ \cite{jonschkowski2020matters}:
\begin{align}
\nonumber \mathcal{L}_{\rm photo}(t,t-1) = & \sum_{u,v}{{\lambda_{\rm L1}  ||\tilde{I_t}(u,v)-I_{t}(u,v)||}_{\rm 1} } \\
& -\lambda_{\rm SSIM} {||\tilde{I_t}(u,v),I_{t}(u,v)||}_{\rm SSIM}.
\label{eq:photometric}
\end{align}
$\mathcal{L}_{\rm photo}(t,t-1)$ is measured by the L1 regularization and structural similarity (SSIM), where $\lambda_{\rm L1}$ and $\lambda_{\rm SSIM}$ correspond to the penalty weights.

Given the objective function, $\mathcal{O}_{\omega}$ can be trained on any pair of consecutive frames. However, minimizing $\sum_{t}\mathcal{L}_{\rm photo}(t,t-1)$ on large-scale datasets will lead to divergence {\color{black}given that} $\mathcal{O}_{\omega}$ is too simple.
To avoid this problem, we propose to train $\mathcal{O}_{\omega}$ on a single image pair, not on datasets composed of numerous image pairs. $\mathcal{O}_{\omega}$ {\color{black}learns} only one sample at each time. In other words, each image pair, whether from training or test datasets, is assigned an independent set of parameters to compute the optical flow. For instance, $O_{t,t-1}$ and $O_{t-1,t-2}$ are estimated by $\mathcal{O}_{\omega_{t, t-1}}(f_{t, t-1}^s)$ and $\mathcal{O}_{\omega_{t-1, t-2}}(f_{t-1, t-2}^s)$, respectively, where the parameters $\omega_{t, t-1} \neq \omega_{t-1, t-2}$. $\omega_{t, t-1}$ and $\omega_{t-1, t-2}$ are optimized by minimizing $\mathcal{L}_{\rm photo}(t,t-1)$ and $\mathcal{L}_{\rm photo}(t-1,t-2)$, respectively.

{\color{black}Notably,} the optical flow is VOS-oriented and we mainly focus on its foreground object regions. Therefore, we decompose the image into foreground and background parts according to the object mask predicted by appearance network. $\mathcal{O}_{\omega}$ is enforced to learn the foreground part of optical flow. {\color{black}With} $\mathcal{O}_{\omega_{t, t-1}}$ {\color{black}as an} example, the optimal parameter $\hat{\omega}_{t,t-1}$ for estimating $O_{t,t-1}$ is obtained by
\begin{equation}
\hat{\omega}_{t,t-1} = \underset{\omega_{t,t-1}}{\arg \min} \{ \mathcal{L}_{\rm photo}(t,t-1; M_t) \},
\label{eq:one-shot}
\end{equation}
\begin{align}
\nonumber \mathcal{L}_{\rm photo}(t,t-1; M_t) =  \sum_{u,v}{({\lambda_{\rm L1}  ||\tilde{I_t}(u,v)-I_{t}(u,v)||}_{\rm 1} } \\
-\lambda_{\rm SSIM} {||\tilde{I_t}(u,v),I_{t}(u,v)||}_{\rm SSIM}) \cdot M_t(u,v),
\label{eq:photometricmask}
\end{align}
where $M_t(u,v) = (P_t(u,v)>0.5)$.
The object mask $M_t$ provides a prior knowledge about the location of foreground moving object. It narrows the solution space of the original objective function $\mathcal{L}_{\rm photo}(t,t-1)$.
In practice, we use a rectangular bounding box to replace $M_t$. The {\color{black}size and position of box} are determined by
{\color{black} the coordinates of $(M_t == 1)$} \cite{bhat2020learning}.

Learning a single optical flow greatly reduces the burden of $\mathcal{O}_{\omega}$. Such a learning scheme is similar to online learning \cite{sahoo2017online}, or one-shot learning \cite{fei2006one}. Learning {\color{black}the foreground parts of optical flow} further releases the pressure. Later in this {\color{black}section}, we will introduce how $\hat{\omega}_{t,t-1}$ is found with a fast speed. For simplicity, $\mathcal{L}_{\rm photo}(t,t-1; M_t)$ is abbreviated as $\mathcal{L}_{\rm p}(\omega_{t, t-1})$.

Minimizing $\mathcal{L}_{\rm p}(\omega_{t, t-1})$ is a nonlinear regression problem, which needs to be solved by iteration.
Given $\omega^{i}_{t,t-1}$ at the $i$th iteration, $\mathcal{L}_{\rm p}(\omega^{i}_{t,t-1})$ can be expanded using the Taylor series, up to the first-order term:
\begin{align}
\label{eq:taylor}
\mathcal{L}_{\rm p}(\omega_{t, t-1}^{i}+\Delta \omega_{t, t-1}^{i}) \approx & \mathcal{L}_{\rm p}(\omega_{t, t-1}^{i}) + \Delta \omega_{t, t-1}^{iT} J_{w_{t,t-1}^i},\\
J_{w_{t,t-1}^i}= & \partial \mathcal{L}_{\rm p}(\omega_{t, t-1}^{i}) / \partial \omega_{t, t-1}^{i},
\end{align}
where $J_{w_{t,t-1}^i}$ is the Jacobian matrix. According to the gradient descent algorithm, $\mathcal{L}_{\rm p}(\omega^{i}_{t,t-1})$ is bound to be descended if $\omega_{t, t-1}^{i}$ is updated along the negative direction of gradient:
\begin{align}
\mathcal{L}_{\rm p}(\omega_{t, t-1}^{i}+\Delta \omega_{t, t-1}^{i}) \approx & \mathcal{L}_{\rm p}(\omega_{t, t-1}^i) - \alpha  ^i J_{w_{t,t-1}^i}^T J_{w_{t,t-1}^i},
\label{eq:descent}
\end{align}
where the increment $\Delta \omega_{t, t-1}^i = -\alpha^i J_{w_{t,t-1}^i}$. $\alpha ^ i$ is the learning rate that determines the stride of $i$th updating. 

$\hat{\omega}_{t,t-1}$ can be found or be approximated by iteratively updating. This process needs to be {\color{black}conducted} quickly when estimating $O_{t,t-1}$ on the fly. Commonly used deep learning optimization algorithms, {\color{black} such as} the SGD \cite{paszke2017pytorch} and Adam \cite{kingma2014adam}, are powerful in updating network parameters. {\color{black}However,} they may take lots of iterations when $\alpha$ is not properly set. For fewer iterations, the RSD optimization algorithm is proposed in this {\color{black}study}, which leads {\color{black}to} a significantly faster convergence speed.

Given $\omega^{i}_{t,t-1}$, the original steepest descent (SD) algorithm finds an optimal increment $\Delta \hat{\omega}^{i}_{t,t-1}$ for $\mathcal{L}_{\rm p}(\omega_{t, t-1}^{i}+\Delta \omega_{t, t-1}^{i})$:
\begin{align}
\Delta \hat{\omega}_{t, t-1}^{i} = \underset{\Delta \omega_{t, t-1}^{i}}{\arg \min} \{ \mathcal{L}_{\rm p}(\omega_{t, t-1}^i + \Delta \omega_{t, t-1}^{i}) \}.
\label{eq:objectiveincrement}
\end{align}
We replace $\Delta \omega^{i}_{t,t-1}$ by $ - \alpha^i J_{w_{t,t-1}^i}$, and {\color{black}we obtain}:
\begin{align}
\hat{\alpha}^i = \underset{\alpha^i}{\arg \min} \{ \mathcal{L}_{\rm p}(\omega_{t, t-1}^i -  \alpha^i J_{w_{t,t-1}^i}) \}.
\label{eq:learingrate}
\end{align}
$\hat{\alpha}^i$ represents the optimal descent stride along the current negative gradient direction. If $\mathcal{L}_{\rm p}(\omega_{t, t-1}^i -  \alpha^i J_{w_{t,t-1}^i})$ is convex and linear, {\color{black}then} $\hat{\alpha}^i$ can be found by solving
\begin{align}
\frac{\partial \mathcal{L}_{\rm p}(\omega_{t, t-1}^i -  \alpha^i J_{w_{t,t-1}^i})}{\partial \alpha^i}=0
\end{align}
However, the requirement of convexity and linearity is too strict for general loss functions in deep learning, {\color{black}which is} the same for $\mathcal{L}_{\rm p}(\omega_{t, t-1}^i -  \alpha^i J_{w_{t,t-1}^i})$. As a result, the minimum {\color{black}cannot} be obtained by simply calculating the stationary point. The proposed RSD algorithm, which is modified from the SD, relaxes this strict requirement. It can be applied to optimize nonconvex or nonlinear objective functions, where the minimum is obtained by a rough estimation.

\begin{figure}[t] \centering
	\begin{subfigure}[b]{0.49\linewidth}
		\centering
		\includegraphics[width=40mm]{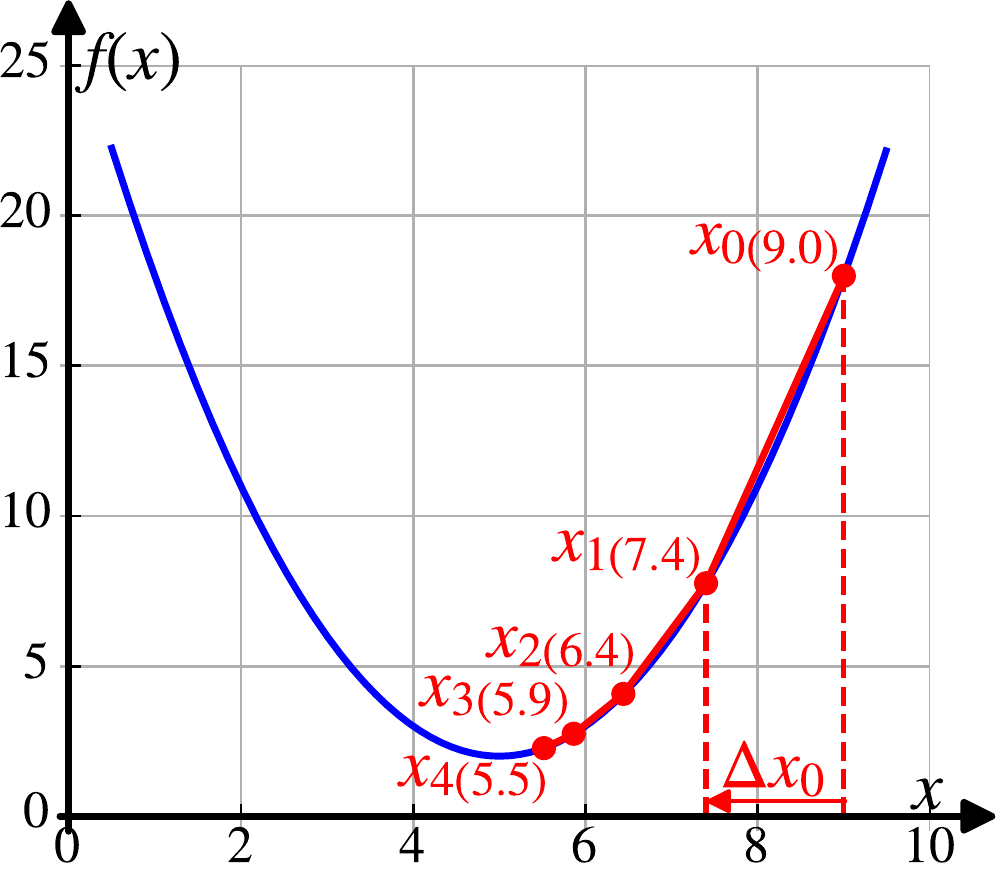}
		\caption{Gradient descent with $\alpha=0.2$}
		\label{fig:optimizationa}
	\end{subfigure} %
	\begin{subfigure}[b]{0.49\linewidth}
		\centering
		\includegraphics[width=40mm]{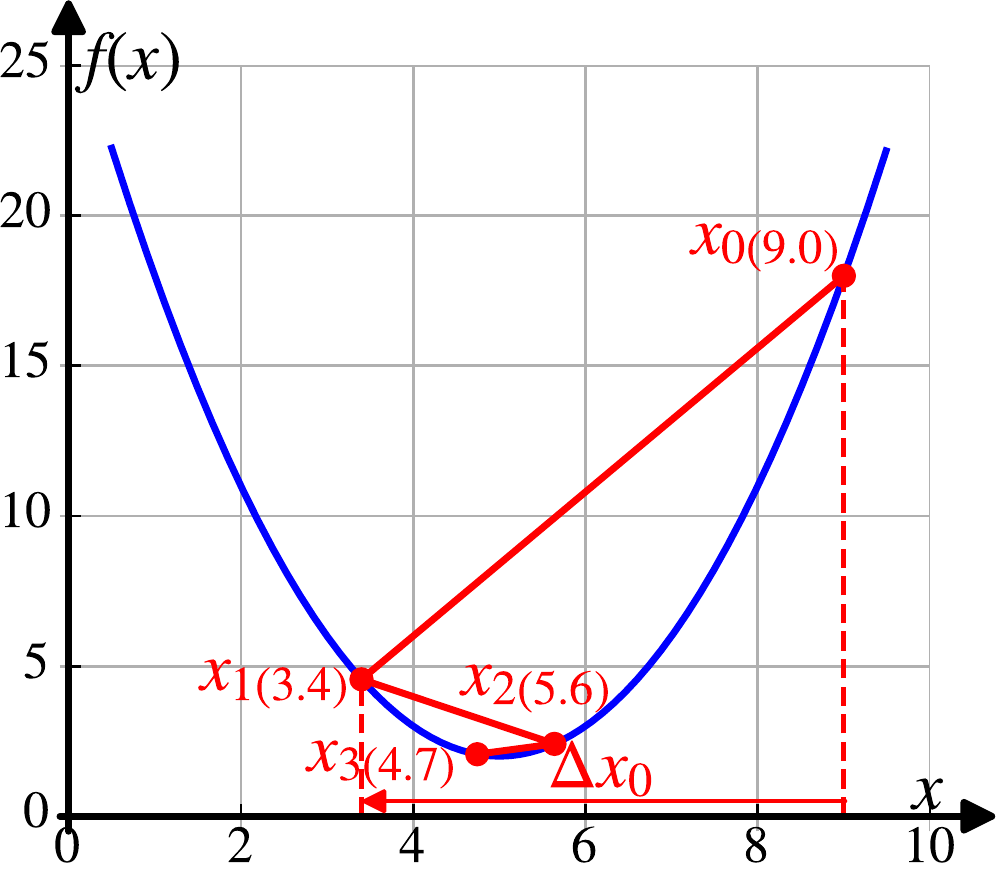}
		\caption{Gradient descent with $\alpha=0.7$}
		\label{fig:optimizationb}
	\end{subfigure}
	
	\begin{subfigure}[b]{0.49\linewidth}
		\centering
		\includegraphics[width=40mm]{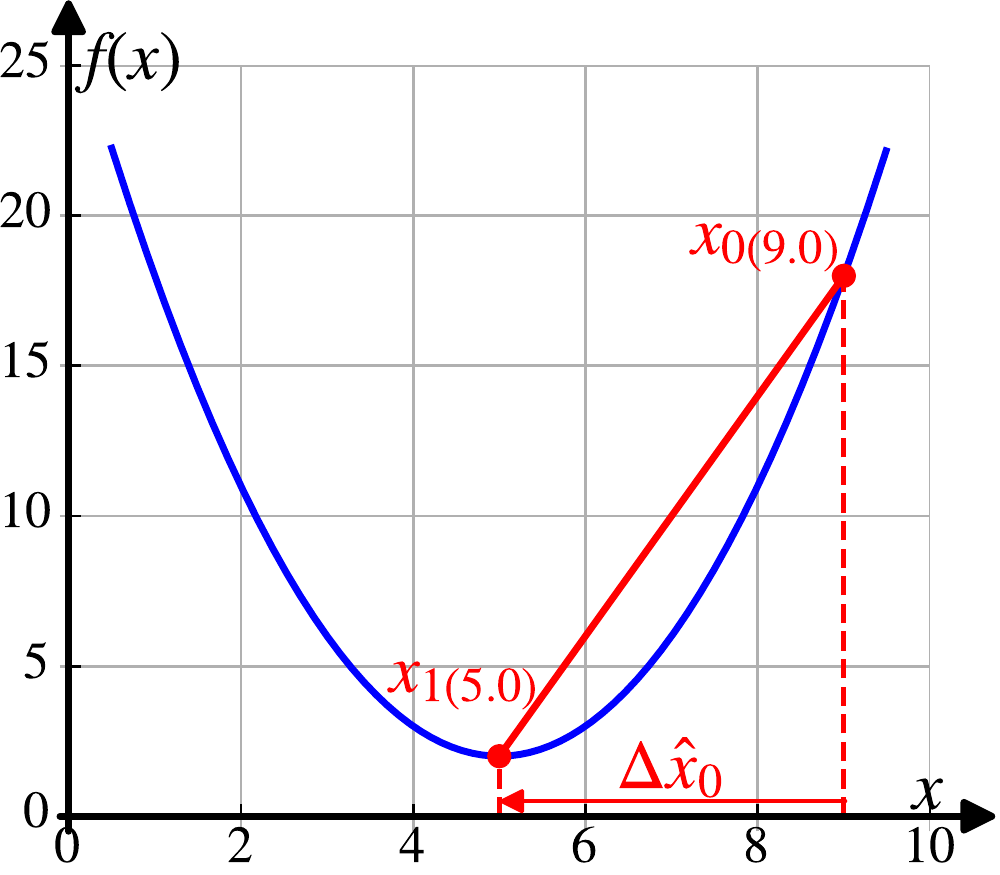}
		\caption{Steepest descent}
		\label{fig:optimizationc}
	\end{subfigure}
	\begin{subfigure}[b]{0.49\linewidth}
		\centering
		\includegraphics[width=40mm]{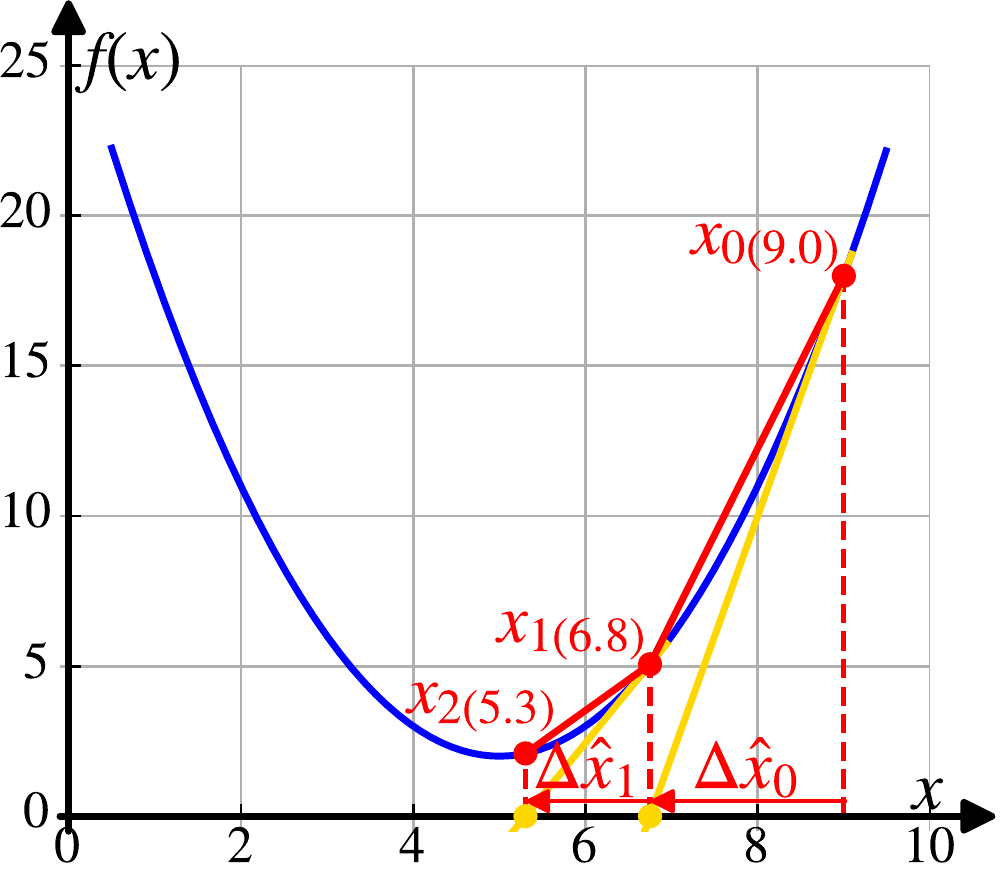}
		\caption{Relaxed steepest descent}
		\label{fig:optimizationd}
	\end{subfigure}
	
	\caption{Comparison of different algorithms for optimizing objective $f(x) = (x-5)^2 + 2$. $f(x)$ is plotted in blue curve. Optimization progress is visualized in red dots and lines. Number in bracket represents the value of current $x_i (i=0,1,2...)$. Yellow line represents $f(x)$'s tangent at $x_i$. {\color{black}Notably,} the optimal solution is given at $\hat{x}=5$.}
	\label{fig:optimization}
\end{figure}

Ideally, the reconstruction error is close to zero when per-pixel correspondence is correctly found. Thus, we use an ideal minimum, 0, to approximately replace the actual minimum of error function $\mathcal{L}_{\rm p}(\omega_{t, t-1}^i -  \alpha^i J_{w_{t,t-1}^i})$. Then, substituting $\hat{\alpha}^i$, we {\color{black}obtain} $\mathcal{L}_{\rm p}(\omega_{t, t-1}^i -  \hat{\alpha}^i J_{w_{t,t-1}^i}) \approx 0$. In other words, at the current $i$th iteration, $w_{t,t-1}^i$ is updated towards an ideal optimal solution, which {\color{black}can} estimate the ground truth optical flow between the foreground object regions of $I_t$ and $I_{t-1}$. {\color{black}Notably,} both the ideal minimum and optimal solution are nonexistent actually, and they are adopted only for approximate computation. According to Equation \ref{eq:descent},
\textcolor{black}{$\hat{\alpha}^i$ and $ \Delta \hat{\omega}^{i}_{t,t-1}$ can be approximated by:}
\begin{align}
\hat{\alpha}^i \approx \frac{\mathcal{L}_{\rm p}(\omega_{t, t-1}^i)}{J_{w_{t,t-1}^i}^T J_{w_{t,t-1}^i}} = \frac{\mathcal{L}_{\rm p}(\omega_{t, t-1}^i)}{{||J_{w_{t,t-1}^i}||}_{\rm F}^2},
\label{eq:alpha}
\end{align}
\begin{align}
\Delta \hat{\omega}^{i}_{t,t-1} = -\hat{\alpha}^i J_{w_{t,t-1}^i} \approx - \mathcal{L}_{\rm p}(\omega_{t, t-1}^i) \frac{J_{w_{t,t-1}^i}}{{||J_{w_{t,t-1}^i}||}_{\rm F}^2},
\label{eq:increment}
\end{align}
where $|| \cdot ||_{\rm F}$ is the Frobenius norm. $\hat{\omega}_{t,t-1}$ can be approximated via only a few iterations rapidly, with $\omega_{t,t-1}$ being updated by an optimal increment at each iteration. Such an optimization algorithm is referred to the RSD. The optical flow estimated by $\hat{\omega}_{t,t-1}$ is {\color{black}sufficiently accurate} for assisting the VOS against the problems of occlusion and object changes.

\begin{figure}[t]
	\centering
	\includegraphics[width=80mm]{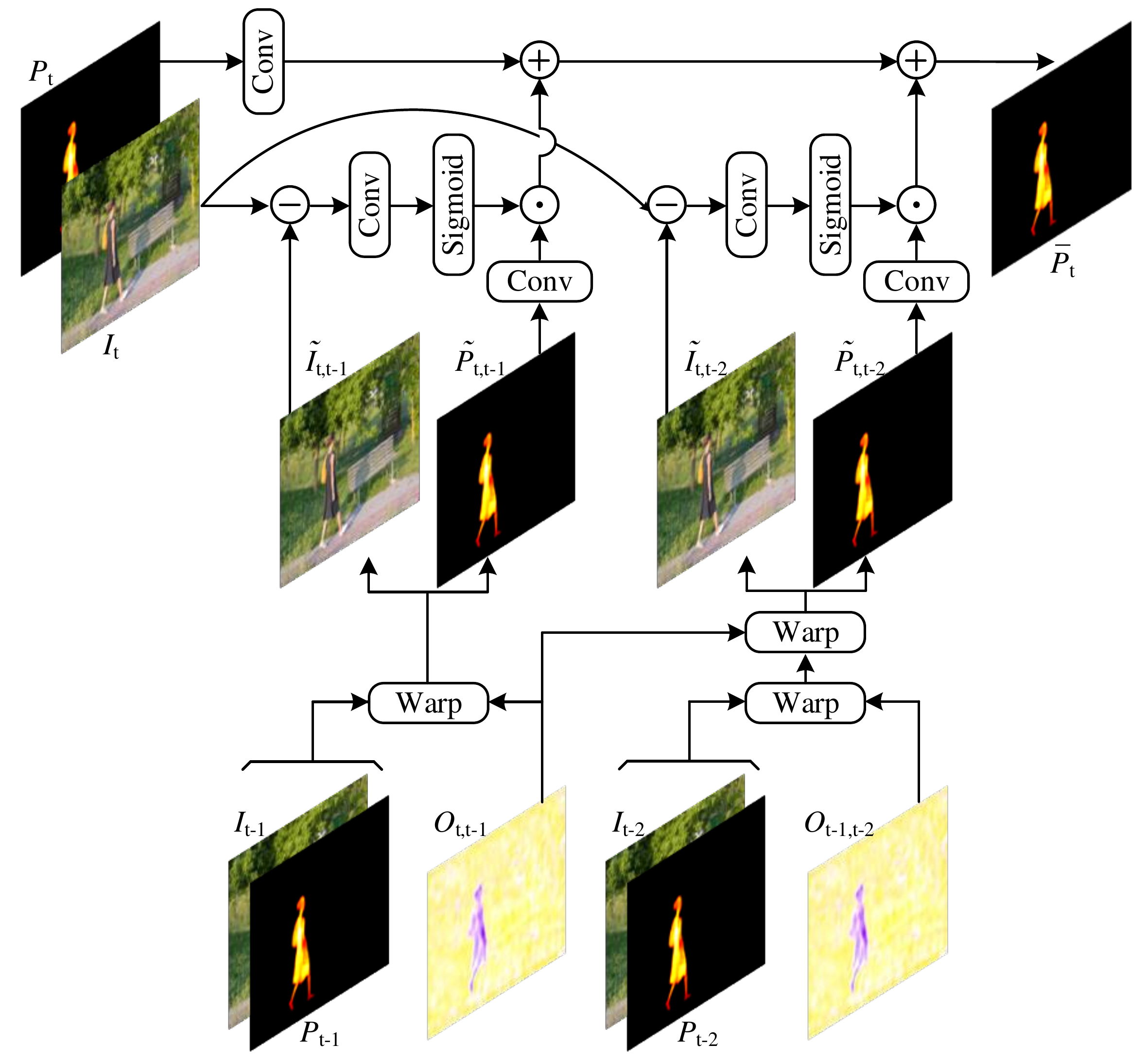}
	\caption{Integration network is implemented by only 5 convolutions, which is lightweight. $\tilde{I}_{t,t-1}$ and $\tilde{P}_{t,t-1}$ indicate the warped $I_{t-1}$ and $P_{t-1}$ by $O_{t,t-1}$. $\tilde{I}_{t,t-2}$ and $\tilde{P}_{t,t-2}$ indicate the warped $I_{t-2}$ and $P_{t-2}$ by $O_{t-1,t-2}$ and  $O_{t,t-1}$.}
	\label{fig:integration}
\end{figure}

{\color{black}We optimize a simple quadratic objective function $f(x)$ as an example, as shown in Figure \ref{fig:optimization}, to have an intuitive comprehension about the RSD algorithm.} 
The gradient descent algorithm can take fewer iterations if $\alpha$ is properly set, {\color{black}such as} 0.7 brings fewer iterations than 0.2.
The SD algorithm can obtain an optimal increment $\Delta \hat{x}_0$ by calculating the stationary point, and {\color{black}it} finds the optimal solution via only one iteration.
{\color{black}Meanwhile,} in the RSD algorithm, the minimum of $f(x)$ is roughly estimated as 0. {\color{black}The assumption is that the stationary point can not be calculated, which is same as in} nonconvex or nonlinear objective functions.
Then, the RSD computes $\hat{\alpha}_i$ and $\Delta \hat{x}_i$ by enforcing $f(x_i + \Delta \hat{x}_i) = 0$ at each iteration. As shown in Figure \ref{fig:optimizationd}, the RSD works well in optimizing $f(x)$, {\color{black}although} $f(x)$'s minimum is not zero.
Intuitively, the RSD {\color{black}finds} the intersection between $f(x)$'s tangent (represented by the yellow line) and $x$-axis, {\color{black}such as} the yellow dot.
The tangent of $f(x)$ at point $(x_0, f(x_0))$ can be described by function $f'(x_0) (x - (x_0 - f(x_0) / f'(x_0)))$, where $f'(x_0) =\partial f(x_0) / \partial x_0$. Obviously, the yellow dot is point $(x_0 - f(x_0) / f'(x_0), 0)$, and the corresponding increment is $-f(x_0) / f'(x_0)$, {\color{black}which is} consistent with Equation \ref{eq:increment}.

\textcolor{black}{The RSD leads to a faster convergence speed than gradient descent-based algorithms while having a wider range of applications than the SD.} $O_{t,t-1}$ is estimated once $\hat{\omega}_{t,t-1}$ is approximated by the RSD. The process of estimating other optical flows is the same {\color{black}as} that of estimating $O_{t,t-1}$.

\subsection{Integration network}
\label{sec:integration}

The integration network refines the initial segmentation result predicted by the appearance network given the optical flow estimated by the motion network and {\color{black}segmentations of previous frames}. It is implemented by a few convolutions, as shown in Figure \ref{fig:integration}. {\color{black}Notably,} the integration network refines object probability maps instead of object masks. For refining $P_t$ of the $t$th frame, $P_{t-1}$ and $P_{t-2}$ of previous frames are utilized. Then, the refined probability map $\bar{P}_t$ is computed by a series of convolutions and non-linearities:

\begin{align}
x_t =& \omega_x * P_t, \\
h_t^{(1)} =& \omega_h^{(1)} * \mathcal{W}(P_{t-1}, O_{t,t-1}), \\
h_t^{(2)} =& \omega_h^{(2)} * \mathcal{W}(\mathcal{W}(P_{t-2}, O_{t-1,t-2}), O_{t,t-1}), \\ \label{eq:forget1}
r_t^{(1)} =& \omega_r^{(1)} *  {|| I_t - \mathcal{W}(I_{t-1}, O_{t,t-1}) ||}_1,\\ \label{eq:forget2}
r_t^{(2)} =& \omega_r^{(2)} *  {|| I_t - \mathcal{W}(\mathcal{W}(I_{t-2}, O_{t-1,t-2}), O_{t,t-1})||}_1,\\ \label{eq:integrate}
\bar{P}_t =& x_t + h_t^{(1)} \cdot (1 - \sigma(r_t^{(1)})) + h_t^{(2)} \cdot (1 - \sigma(r_t^{(2)})).
\end{align}
$\bar{P}_t$ {\color{black}depends} on two-stream representations $x_t$ and $h_t$. $x_t$ is the linear transformation of $P_t$. $h_t$ is composed of $h_t^{(1)}$ and $h_t^{(2)}$, which are the linear transformations of calibrated $P_{t-1}$ and $P_{t-2}$ by optical flows.

Different from simply propagating previous predictions in \cite{luiten2018premvos}, the integration network further includes $r_t^{(1)}$ and $r_t^{(2)}$ to control how much of history information is let through into the current result.
As indicated by Equations \ref{eq:forget1} and \ref{eq:forget2}, $r_t$ is computed according to the reconstruction error ${||\tilde{I}_t - I_t||}_1$. A small error at $(u, v)$ means that pixel correspondence is accurate with high confidence. Then, $(1 - \sigma(r_t(u,v)))$ is tend to give a larger value to let $h_t(u,v)$ be passed into $\bar{P}_t$. On the contrary, a large reconstruction error indicates that the optical flow here is inaccurate. $h_t(u,v)$ may be disturbed by a mistaken warping. Thus, we {\color{black}do not} want this unsafe information to be passed, where we use $r_t(u,v)$ to suppress the passing.

The integration network segments objects in continuous video frames by memorizing {\color{black}static appearances and dynamic motions of objects}. {\color{black}Notably,} $\omega_h^{(1)}$ and $\omega_h^{(2)}$ {\color{black}cannot} be shared {\color{black}given that} $P_{t-1}$ and $P_{t-2}$ have different impacts on $\bar{P}_{t}$. The most recent frame should be prioritized in the current prediction. {\color{black}Exploiting two previous frames' information is sufficient} to produce a temporal-consistent prediction. Adding more frames, such as $h_t^{(3)}$ transformed from $P_{t-3}$, {\color{black}does not} bring more gains. Compared with existing optical flow-based VOS methods \cite{cheng2017segflow, jain2017fusionseg, tokmakov2017learning, perazzi2017learning}, the proposed integration network works in a more simple and explainable way.

\section{Experiments}
Extensive experiments are conducted to demonstrate the performance of our proposed FAMINet. Before that, we evaluate the quality of optical flow estimated by the motion network {\color{black}given that} the optical flow plays an important role in our method.

\subsection{Datasets and Metrics}
\subsubsection{Datasets}
For evaluating the performance of VOS methods, two VOS benchmarks, DAVIS \cite{perazzi2016benchmark} and Youtube-VOS \cite{xu2018youtube}, are utilized. DAVIS is composed of DAVIS 2016 \cite{perazzi2016benchmark} and 2017 \cite{pont20172017}. DAVIS 2016 only contains annotated masks for individual objects, while DAVIS 2017 is a multi-object extension of 2016. We use DAVIS 2017 training set with 60 videos for training, and DAVIS 2016 validation set with 20 videos, DAVIS 2017 validation set with 30 videos, and DAVIS 2017 test split with 30 videos for evaluation. Youtube-VOS is a large-scale dataset annotated with multiple objects. We use 3471 videos from 65 categories for training, and 474 videos with additional 26 unseen categories for evaluation.

For evaluating the performance of optical flow methods, DAVIS 2017 validation set is utilized.

\subsubsection{Metrics}
For measuring the segmentation accuracy, the average of $\mathcal{J}$ score and $\mathcal{F}$ score are computed. $\mathcal{J}$ score is defined as the intersection-over-union (IoU) between predicted segmentation mask and ground truth segmentation mask. $\mathcal{F}$ score is defined as the boundary similarity between the boundary of prediction and ground truth. $\mathcal{J}$ and $\mathcal{F}$ are reported in percent, and \% is omitted for simplicity. We report $\mathcal{J}$\&$\mathcal{F}$ in some cases, as the average of $\mathcal{J}$ and $\mathcal{F}$.

For measuring the accuracy of optical flow, the {\color{black}peak signal-to-noise ratio} (PSNR) and {\color{black}structural similarity} (SSIM) are computed between the {\color{black}target and reconstructed images} warped from source image. We also warp {\color{black}the ground truth segmentation mask of the source image} by optical flow, and measure the $\mathcal{J}$ and $\mathcal{F}$ between the warped segmentation mask and {\color{black}ground truth segmentation mask of target image}.

Moreover, we use the frames per second (FPS) for measuring the {\color{black}inference speed of algorithm}. {\color{black}Higher $\mathcal{J}$, $\mathcal{F}$, PSNR, SSIM, and FPS are better.}
\subsection{Implementing details}
We use Pytorch based on Python to implement all methods. Experiments are conducted on the 4029GP-TRT server carried with NVIDIA TITAN RTX GPU of 24GB memory. We give more details about network implementing, training, and test.
\subsubsection{Network details}
\label{details:network}
\begin{itemize}
	\item \textbf{Feature extractor}: {\color{black}It} is implemented by ResNet18 \cite{he2016deep}, pre-trained on ImageNet \cite{russakovsky2015imagenet}. Its parameters are kept {\it{fixed}} during training and test.
	\item \textbf{Appearance network}: {\color{black}It} is implemented by the FRTM-fast in \cite{robinson2020learning}. We choose the FRTM-fast because it achieves a high-accuracy segmentation result with a fast inference speed, compared with existing VOS methods.
	\item \textbf{Motion network}: {\color{black}It} is implemented by two convolutions. Note that input is the feature map from block {\it{conv\_3x}} in ResNet18. Thus, $f_{t, t-1}^s$ in Equation \ref{eq:opticalflow}, where $s=8$, has 256 feature channels. $\omega^{(1)}$ is $256 \times 96$ kernels with the size of $1 \times 1$, and $\omega^{(2)}$ is $96 \times 2$ kernels with the size of $3 \times 3$. We up-sample output with a scale of 8 by {\it{bilinear}} interpolation to match the size of image. The two convolutions are optimized by RSD algorithm on the fly.
	\item \textbf{Integration network}: {\color{black}The} kernel size of all convolutional layers is $3 \times 3$. $\omega_r$ contains $3 \times 1$ kernels, while $\omega_x$ and $\omega_h$ contain only 1 kernel.

\end{itemize}

\begin{table}[t]
\caption{Sanity check for optical flow on DAVIS 2017 validation set. Our proposed algorithm is {\bf{shown in bold}}. We crop the target and reconstructed images by object bounding box, and then calculate the PSNR and SSIM between cropped images.}
\label{scflow}
    \centering
    \scriptsize
	\setlength\tabcolsep{3pt} 
    \begin{tabular}{l|c c c c|ccccl}
    \toprule
        \textbf{algorithm} & \bm{$M$} & \textbf{share} & \bm{$\alpha$} & \textbf{iterations} & \textbf{PSNR} & \textbf{SSIM} & \bm{$\mathcal{J}$} & \bm{$\mathcal{F}$} & \textbf{FPS} \\ \midrule
        w/o warp & - & - & - & - & 17.47 & 0.4947 & 71.6 & 84.0 & - \\ \midrule
        RAFT \cite{teed2020raft} & - & - & - & - & 21.50 & 0.7460 & 78.2 & 89.7 & 7.93 \\
        RAFT-DV \cite{teed2020raft} & - & - & - & - & 23.87 & 0.7899 & 84.7 & 94.2 & 7.93 \\ \midrule
        Motion-DV & - & - & - & - & 18.87 & 0.5480 & 76.4 & 88.9 & 219.19 \\ \midrule
        Motion-SGD & \xmark & \cmark & 1e-4 & 1 & 17.46 & 0.4932 & 68.1 & 80.0 & 127.59 \\
        Motion-SGD & \xmark & \cmark & 1e-5 & 1 & 17.76 & 0.5215 & 70.7 & 82.9 & 127.59 \\
        Motion-SGD & \xmark & \cmark & 1e-6 & 1 & 17.63 & 0.5059 & 71.6 & 84.0 & 127.59 \\ \midrule
        Motion-SGD & \cmark & \cmark & 1e-4 & 1 & 17.85 & 0.5129 & 70.7 & 82.8 & 123.47 \\
        Motion-SGD & \cmark & \cmark & 1e-5 & 1 & 17.87 & 0.5273 & 71.0 & 83.2 & 123.47 \\
        Motion-SGD & \cmark & \cmark & 1e-6 & 1 & 17.66 & 0.5077 & 71.7 & 84.0 & 123.47 \\ \midrule
        Motion-SGD & \cmark & \xmark & 1e-3 & 1 & 17.41 & 0.4904 & 66.2 & 77.4 & 123.47 \\
        Motion-SGD & \cmark & \xmark & 1e-4 & 1 & 18.29 & 0.5427 & 73.1 & 85.4 & 123.47 \\
        Motion-SGD & \cmark & \xmark & 1e-5 & 1 & 17.73 & 0.5109 & 71.7 & 84.1 & 123.47 \\ \midrule
        Motion-SGD & \cmark & \xmark & 1e-4 & 3 & 19.09 & 0.5869 & 75.4 & 87.5 & 66.23 \\
        Motion-SGD & \cmark & \xmark & 1e-4 & 5 & 19.60 & 0.6129 & 76.9 & 88.8 & 46.81 \\
        Motion-SGD & \cmark & \xmark & 1e-4 & 10 & 20.31 & 0.6452 & 78.5 & 90.3 & 26.35 \\
        Motion-SGD & \cmark & \xmark & 1e-4 & 20 & 22.21 & 0.7025 & 82.8 & 93.2 & 13.92 \\ \midrule
        Motion-Adam & \cmark & \xmark & 1e-3 & 1 & 17.38 & 0.4884 & 69.9 & 82.1 & 118.07 \\
        Motion-Adam & \cmark & \xmark & 1e-4 & 1 & 17.61 & 0.5049 & 71.6 & 84.0 & 118.07 \\
        Motion-Adam & \cmark & \xmark & 1e-5 & 1 & 17.60 & 0.5040 & 71.6 & 84.0 & 118.07 \\
        Motion-Adam & \cmark & \xmark & 1e-4 & 3 & 17.58 & 0.5006 & 71.3 & 83.6 & 66.70 \\
        Motion-Adam & \cmark & \xmark & 1e-4 & 5 & 17.45 & 0.4929 & 70.6 & 82.5 & 44.25 \\ \midrule
        \bf{Motion}\bf{-RSD} & \xmark & \cmark & - & 1 & 17.90 & 0.5338 & 67.6 & 79.4 & 121.33 \\
        \bf{Motion}\bf{-RSD} & \cmark & \cmark & - & 1 & 19.17 & 0.5770 & 74.5 & 87.8 & 121.33 \\
        \bf{Motion}\bf{-RSD} & \cmark & \xmark & - & 1 & 20.08 & 0.6015 & 79.3 & 91.6 & 121.33 \\
        \bf{Motion}\bf{-RSD} & \cmark & \xmark & - & 3 & 20.73 & 0.6301 & 81.0 & 92.8 & 65.06 \\
        \bf{Motion}\bf{-RSD} & \cmark & \xmark & - & 5 & 21.24 & 0.6575 & 81.8 & 93.2 & 44.60 \\ \bottomrule
    \end{tabular}
\end{table}

\subsubsection{Training details}
The overall training process is divided into two stages. In the first stage, the appearance network is trained independently following the training scheme in \cite{robinson2020learning}. Each training sample in a batch includes 3 image frames from a sequence, with 1 frame for initializing the target model and other 2 frames for training. The loss on each sample is formalized by binary cross entropy (BCE):
\begin{align}
\mathcal{L}_{\rm stage1} =\sum_{t=1}^2 {||M_t, Y_t||}_{\rm BCE}.
\end{align}

In the second stage, the appearance and integration networks are trained jointly given the optical flow estimated by the motion network. \textcolor{black}{$\lambda_{\rm L1}$ and $\lambda_{\rm SSIM}$ in Equation \ref{eq:photometricmask} are empirically set as 0.15 and 0.85, respectively, according to \cite{zhao2015loss, yin2018geonet, jonschkowski2020matters}.} Each training sample includes 4 frames, with 1 frame for initializing and other 3 {\it{consecutive}} frames for training. The loss on each sample is given by:
\begin{align}
\mathcal{L}_{\rm stage2} =\sum_{t=1}^2 {||M_t, Y_t||}_{\rm BCE} + {||\bar{M}_3, Y_3||}_{\rm BCE},
\end{align}
where $M_1$ and $M_2$ are from the appearance network, and refined $\bar{M}_3$ is from the integration network. The batch size is 16. The initial learning rate is 2e-4 and decreased by a factor of 0.5 every 10 epochs, where the total training epochs in the second stage is 50. The optimizer is Adam \cite{kingma2014adam}.
\subsubsection{Test details}
During test, parameters of $\mathcal{D}_{\theta}$ and the integration network are kept fixed. Each target object in a sequence is assigned an independent $\mathcal{T}_{\tau}$ and $\mathcal{O}_{\omega}$. $\mathcal{T}_{\tau}$ is updated as in \cite{robinson2020learning}. $\mathcal{O}_{\omega}$ is updated as described in Section \ref{sec:selection}.

\subsection{Sanity check for optical flow}
\label{sec:sanity}
We conduct a sanity check for optical flow on DAVIS 2017 validation set, to ensure that the optical flow can be estimated accurately and efficiently. {\color{black}Notably,} we let the source and target images be the previous and current images in a pair of adjacent video frames, respectively. The reconstructed image is warped from source image through the optical flow. The performances of optical flow methods are shown in Table \ref{scflow}.

\begin{figure}[t] \centering
	\begin{subfigure}[b]{0.49\linewidth}
		\centering
		\includegraphics[width=40mm]{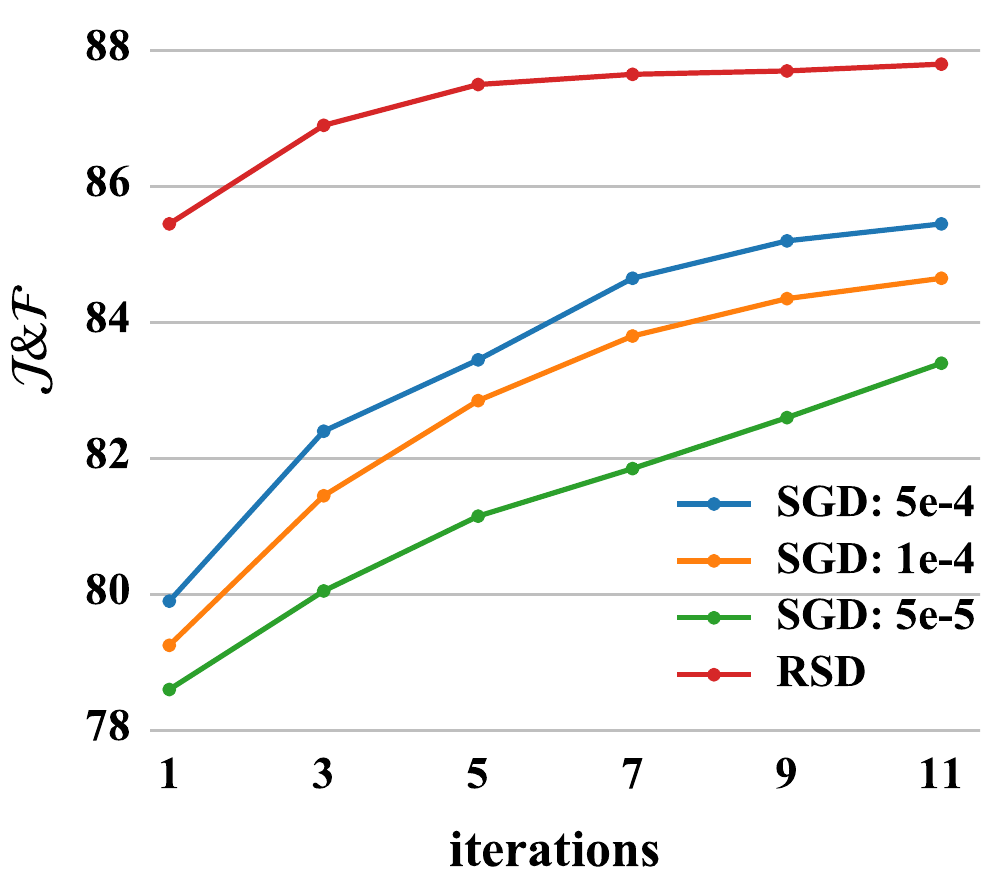}
		\caption{$\mathcal{J}$\&$\mathcal{F}$ of RSD and SGD with different learning rates.}
		\label{fig:rsd-itr}
	\end{subfigure} %
	\begin{subfigure}[b]{0.49\linewidth}
		\centering
		\includegraphics[width=40mm]{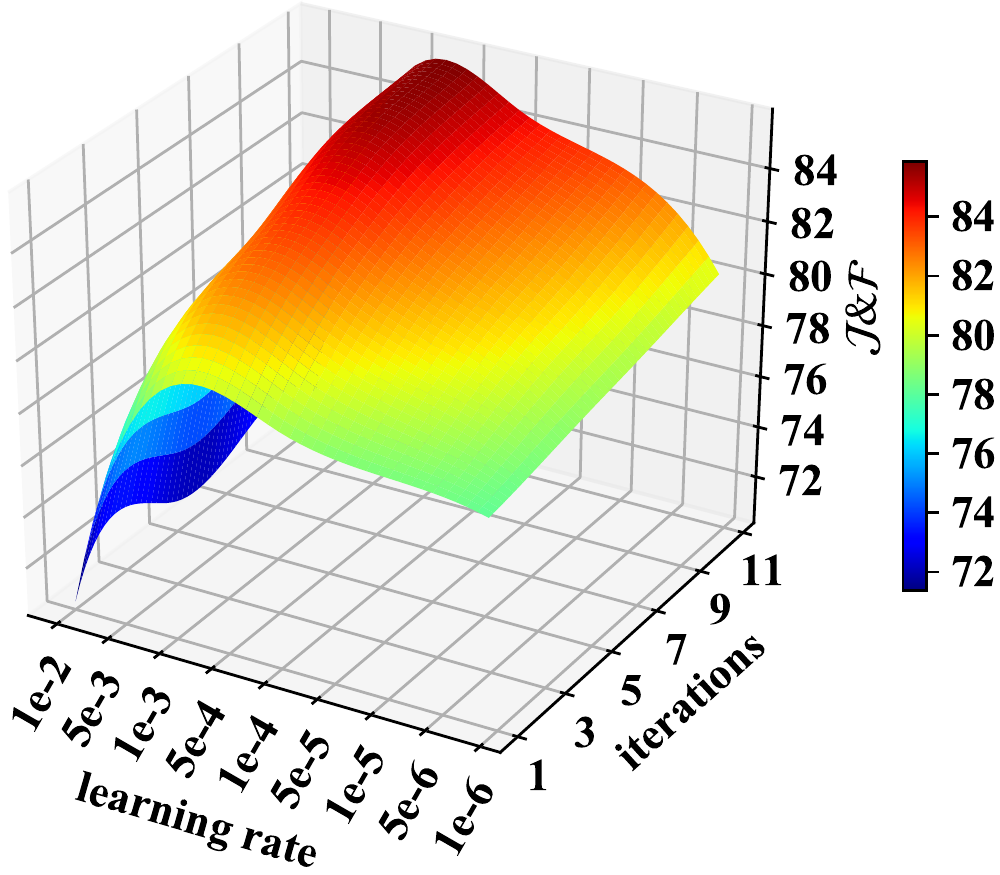}
		\caption{$\mathcal{J}$\&$\mathcal{F}$ of SGD with different learning rates.}
		\label{fig:sgd-itr}
	\end{subfigure}
	
	\caption{Comparison of the optical flows estimated by the Motion-RSD (abbreviated as RSD) and Motion-SGD (abbreviated as SGD) on DAVIS 2017 validation set. $\mathcal{J}$\&$\mathcal{F}$ is utilized as the evaluation metric {\color{black}given that} we mainly focus on optical flow's ability of registering object masks. The results of RSD and SGD ranging from 1 to 11 iterations are drawn.}
\end{figure}

Different optical flow methods are listed on the left, and each of them is explained as {\color{black}follows}:
\begin{itemize}
	\item {w/o warp}: {\color{black}It} indicates that the previous source image/object mask is directly {\color{black}regarded} as the reconstructed image/object mask without warping by the optical flow.
	\item {RAFT}: {\color{black}It} indicates the state-of-the-art optical flow method proposed in \cite{teed2020raft}.
	\item {RAFT-DV}: it indicates the RAFT trained on DAVIS 2017 training set by the unsupervised learning \cite{jonschkowski2020matters} offline.
	\item {Motion-DV}: {\color{black}It} contains a ResNet18, two convolutional layrs $\mathcal{O}_{\omega}$, and an up-sampling layer, as described in Section \ref{details:network}. $\mathcal{O}_{\omega}$ is trained on DAVIS 2017 training set offline. ResNet18 is kept {\it{fixed}}.
	\item {Motion-SGD, Motion-Adam {\color{black}and} Motion-RSD}: {\color{black}They} are all implemented by the same network with Motion-DV. $\mathcal{O}_{\omega}$ is optimized by the SGD, Adam, and proposed RSD algorithms online, respectively, as described in Section \ref{sec:motion}. ResNet18 is kept {\it{fixed}}.

\end{itemize}

\begin{figure*}[t]
	\centering
	\includegraphics[width=170mm]{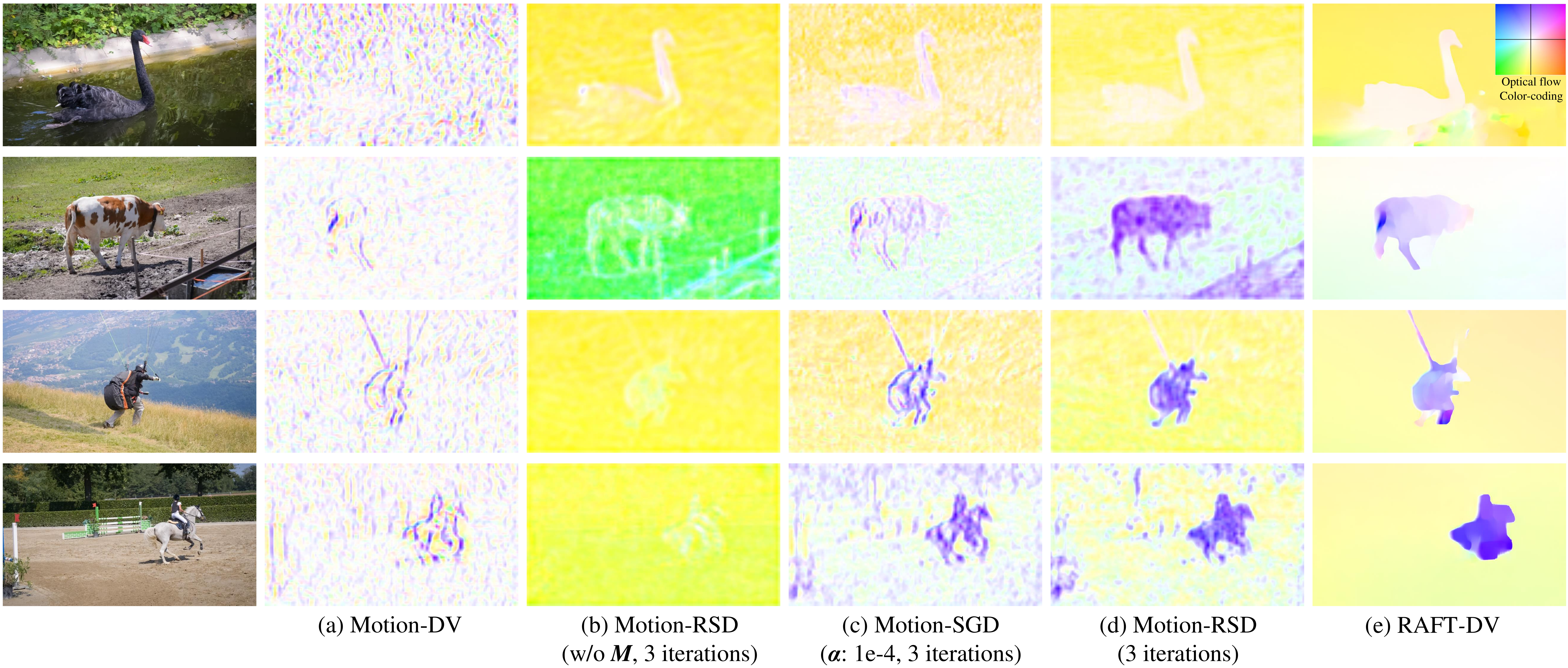}
	\caption{\textcolor{black}{Visualization of the magnitude of the optical flows estimated by different algorithms. The optical flows are color-coded by \cite{baker2011database}, as shown in the upper right corner. Visualization examples are from DAVIS dataset. Foreground optical flow estimated by the Motion-RSD is roughly consistent with that estimated by the RAFT-DV.}}
	\label{fig:flow-compare}
\end{figure*}

\begin{figure}[t]
	\centering
	\includegraphics[width=85mm]{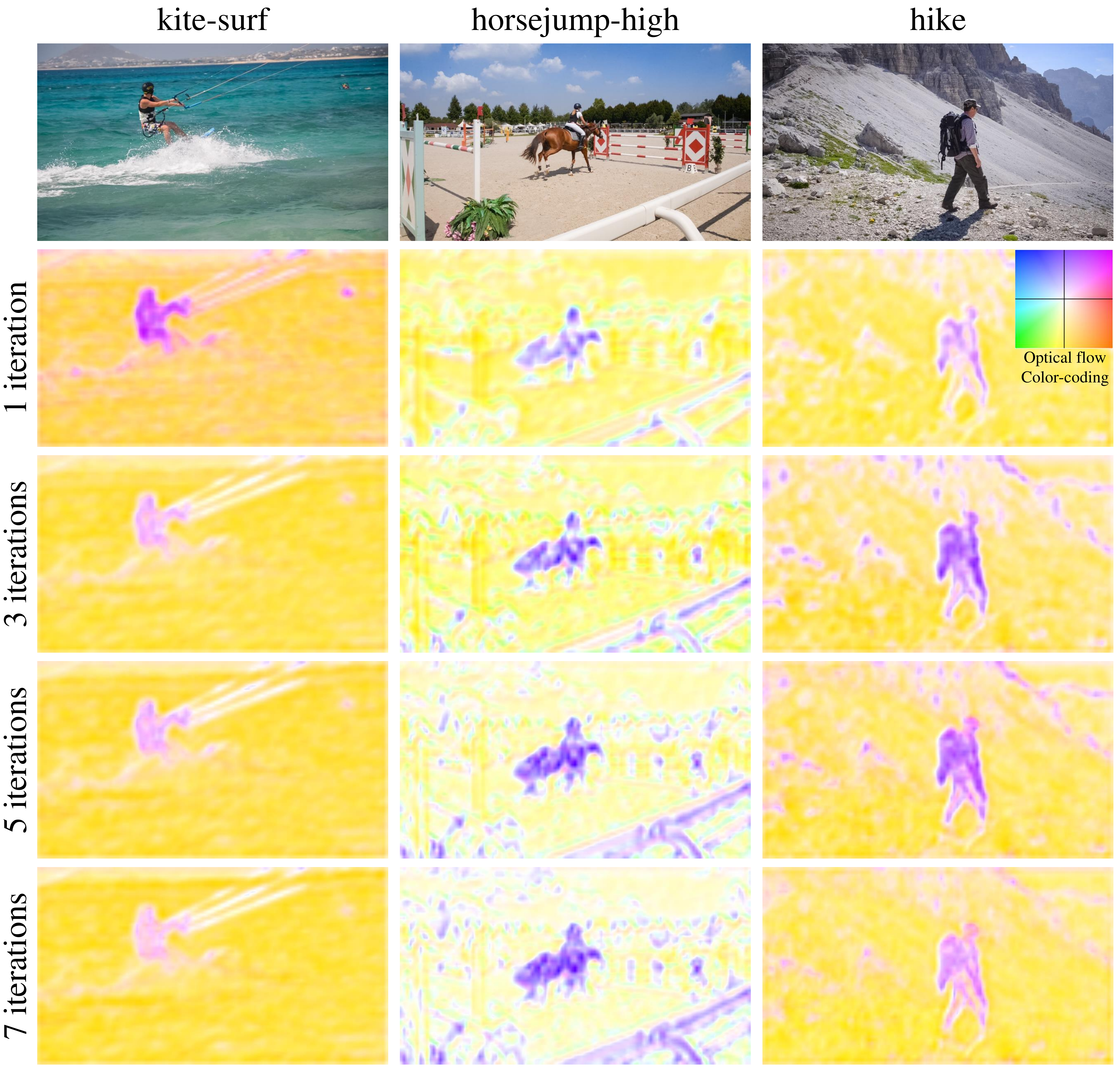}
	\caption{\textcolor{black}{Visualization of the magnitude of the optical flows estimated by the Motion-RSD with different iterations. The optical flows are color-coded by \cite{baker2011database}, as shown in the upper right corner. Visualization examples are from DAVIS dataset.}}
	\label{fig:flow-update}
\end{figure}

Different experimental conditions are listed on the middle, and each of them is explained as {\color{black}follows}:
\begin{itemize}
	\item \bm{$M$}: \cmark \ indicates that Objective \ref{eq:photometricmask} is utilized for optimizing $\mathcal{O}_{\omega}$ online, while \xmark \ indicates that Objective \ref{eq:photometric} without object mask is utilized. The object mask is the ground truth object mask of previous source image. It is also padded with 20 pixels against large displacements.
	\item {\bf{share}}: \cmark \ indicates that the parameters of $\mathcal{O}_{\omega}$ is shared when estimating optical flows of different moving objects. \xmark \ indicates that each object in a sequence is assigned an independent $\mathcal{O}_{\omega}$, {\color{black}which is the same as} the target model $\mathcal{T}_{\tau}$. 
	\item \bm{$\alpha$}: The learning rate adopted for optimizing $\mathcal{O}_{\omega}$ online. For the SGD and Adam, the learning rate is needed to be preset manually. For our proposed RSD, the learning rate is calculated automatically.
	\item {\bf{iterations}}: The number of iterations performed when optimizing $\mathcal{O}_{\omega}$ on a single image pair at each time.

\end{itemize}

As shown on the first row of Table \ref{scflow}, previous and current ground truth object masks {\color{black}have} a high IoU (up to 71.6\%), {\color{black}which implies} that huge potential temporal information can be exploited for the VOS. Reconstruction errors are greatly reduced after warping previous image by the {\color{black}optical flow of RAFT}.
The RAFT-DV further improves $\mathcal{J}$ and $\mathcal{F}$ from the RAFT, {\color{black}which implies} that unsupervised learning optical flow is important for registering previous object mask to the current.

{\color{black}As indicated by the results of Motion-SGD, $\mathcal{O}_{\omega}$ is optimized to produce positive optical flow, with learning rate set as 1e-4, by minimizing Objective \ref{eq:photometricmask} and non-sharing parameters across objects.} Therefore, the burden of learning optical flow is greatly mitigated by decomposing image into foreground and background parts and by learning motions of a single object. The Motion-SGD performs better with increasing iterations, and {\color{black}it} surpasses the Motion-DV over 20 iterations. This {\color{black}observation} demonstrates that a lightweight network is more suitable for being optimized on a few samples online, than being trained on large-scale datasets offline.

The Motion-Adam performs poorly no matter what learning rate or number of iterations is adopted. The optimizer Adam may not apply to small-scale datasets or networks \cite{kingma2014adam}.

The Motion-RSD with only 1 iteration surpasses the Motion-SGD with 10 iterations in $\mathcal{J}$ and $\mathcal{F}$. Clearly, the RSD algorithm leads {\color{black}to} a significantly faster convergence speed than the SGD. Compared {\color{black}with} the RAFT-DV, the Motion-RSD sacrifices 2.9 to 5.4 percent in $\mathcal{J}$, but {\color{black}it} achieves a 5.6 to 15.3 times faster inference speed. The Motion-RSD achieves a good trade-off between accuracy and efficiency.

\begin{algorithm}[t]
\caption{Original RSD algorithm in motion network}
\label{alg:A}
\begin{algorithmic}
\STATE Given a video sequence $\{I_0, I_1, I_2, \dots, I_T  \}$.
\STATE {\bf{for}} $t=0$ $\rightarrow$ $t=T$  {\bf{do}}:
\STATE \ \ \ obtain the $t$th frame $I_t$ and $f_t^8$ by $\mathcal{F}_\theta$;
\STATE \ \ \ {\bf{if}} $t = 0$:
\STATE \ \ \ \ \ \ cache $I_t$ and $f_t^8$ into CPU memory;
\STATE \ \ \ {\bf{else}}:
\STATE \ \ \ \ \ \ index $I_{t-1}$ and $f_{t-1}^8$ from CPU memory;
\STATE \ \ \ \ \ \ obtain $f_{t,t-1}^8$ by concatenating $f_{t}^8$ and $f_{t-1}^8$;
\STATE \ \ \ \ \ \ optimize $\mathcal{O}_{\omega}$ by RSD given $f_{t,t-1}^8$, $I_t$, and $I_{t-1}$; 
\STATE \ \ \ \ \ \ estimate optical flow by $\mathcal{O}_{\omega}$ given $f_{t,t-1}^8$;
\STATE \ \ \ \ \ \ cache $I_t$ and $f_t^8$ into CPU memory.
\STATE \ \ \ {\bf{end if}}
\STATE {\bf{end for}}
\end{algorithmic}
\end{algorithm}

\begin{algorithm}[t]
\caption{Modified RSD algorithm in motion network}
\label{alg:B}
\begin{algorithmic}
\STATE Given a video sequence $\{I_0, I_1, I_2, \dots, I_T  \}$.
\STATE Given an update interval $N$.
\STATE {\bf{for}} $t=0$ $\rightarrow$ $t=T$  {\bf{do}}:
\STATE \ \ \ obtain the $t$th frame $I_t$;
\STATE \ \ \ obtain $f_t^8$ by $\mathcal{F}_\theta$ and $M_t$ by appearance network;
\STATE \ \ \ {\bf{if}} $t = 0$ {\bf{or}} $t \ {\rm mod} \ N \ != 0$:
\STATE \ \ \ \ \ \ cache $I_t$ and $f_t^8$ into CPU memory;
\STATE \ \ \ {\bf{else}}:
\STATE \ \ \ \ \ \ initialize 3 batches, \bm{$f_{c,p}^8$}, \bm{$I_{c}$}, and \bm{$I_{p}$}
\STATE \ \ \ \ \ \ {\bf{for}} $i=1$ $\rightarrow$ $i=N$  {\bf{do}}:
\STATE \ \ \ \ \ \ \ \ \ index $I_{t-i}$ and $f_{t-i}^8$ from CPU memory;
\STATE \ \ \ \ \ \ \ \ \ obtain $f_{t-i+1,t-i}^8$ by concatenating $f_{t-i+1}^8$ and
\STATE \ \ \ \ \ \ \ \ \ $f_{t-i}^8$, and put $f_{t-i+1,t-i}^8$ into batch \bm{$f_{c,p}^8$};
\STATE \ \ \ \ \ \ \ \ \ put $I_{t-i+1}$ into batch \bm{$I_{c}$}, and $I_{t-i}$ into batch \bm{$I_{p}$};
\STATE \ \ \ \ \ \ {\bf{end for}}
\STATE \ \ \ \ \ \ check if the batch size of \bm{$f_{c,p}^8$}, \bm{$I_{c}$}, and \bm{$I_{p}$} equals $N$;
\STATE \ \ \ \ \ \ optimize $\mathcal{O}_{\omega}$ by RSD given \bm{$f_{c,p}^8$}, \bm{$I_{c}$}, and \bm{$I_{p}$};
\STATE \ \ \ \ \ \ estimate a batch of optical flows by $\mathcal{O}_{\omega}$ given \bm{$f_{c,p}^8$};
\STATE \ \ \ \ \ \ cache $I_t$ and $f_t^8$ into CPU memory.
\STATE \ \ \ {\bf{end if}}
\STATE {\bf{end for}}
\end{algorithmic}
\end{algorithm}

Visualization of estimated optical flows is shown in Figure \ref{fig:flow-compare}. The outputs of the Motion-DV are in a mess, {\color{black}which shows} that the lightweight $\mathcal{O}_{\omega}$ is divergent after being trained offline. Comparing Figures \ref{fig:flow-compare}b and \ref{fig:flow-compare}d, we can see that $\mathcal{O}_{\omega}$ {\color{black}can} focus on the most important foreground object motions when minimizing Objective \ref{eq:photometricmask} {\color{black}instead} of Objective \ref{eq:photometric}. Moreover, the optical flow estimated by the Motion-RSD is more consistent and has clearer object boundaries than that by the Motion-SGD. The RAFT-DV produces the most accurate optical flow.

We conduct more experiments on selecting different learning rates of the Motion-SGD. The results are summarized in Figure \ref{fig:rsd-itr}. The Motion-RSD has significant advantage over the Motion-SGD in estimating optical flow no matter what learning rate is adopted in the Motion-SGD. Performance gap between the two algorithms is more obvious within fewer iterations. To be more convincing, the performance of Motion-SGD with learning rate ranging from 1e-6 to 1e-2 are evaluated thoroughly. The results are plotted in Figure \ref{fig:sgd-itr}, with x-axis corresponding to learning rate, y-axis to iterations, and z-axis to $\mathcal{J}$\&$\mathcal{F}$. The highest $\mathcal{J}$\&$\mathcal{F}$ achieved by the Motion-SGD is still lower than that by the Motion-RSD.

\begin{figure}[t] \centering
	\begin{subfigure}[b]{0.98\linewidth}
		\centering
		\includegraphics[width=80mm]{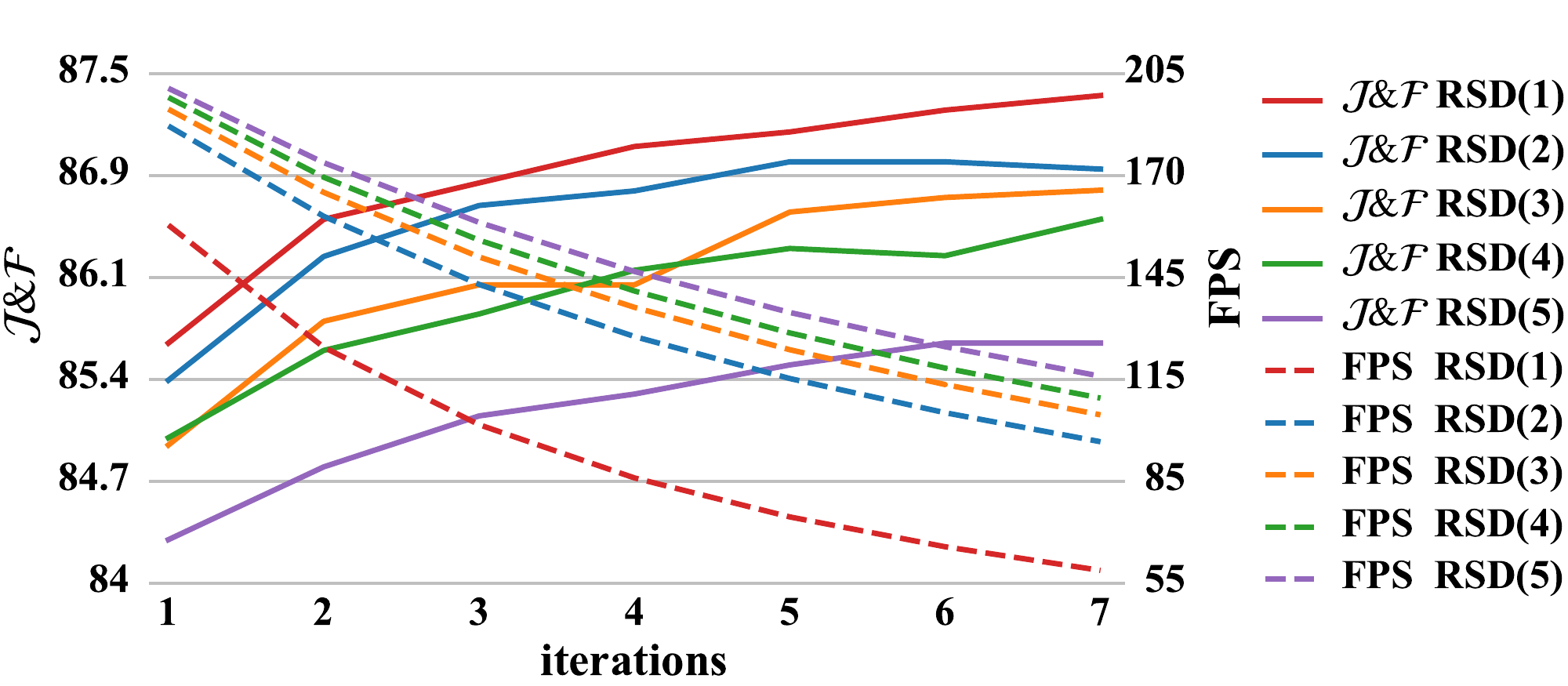}
		\caption{Comparison on DAVIS 2016 validation set}
		\label{fig:rsd-dv16}
	\end{subfigure} %
		\begin{subfigure}[b]{0.98\linewidth}
		\centering
		\includegraphics[width=80mm]{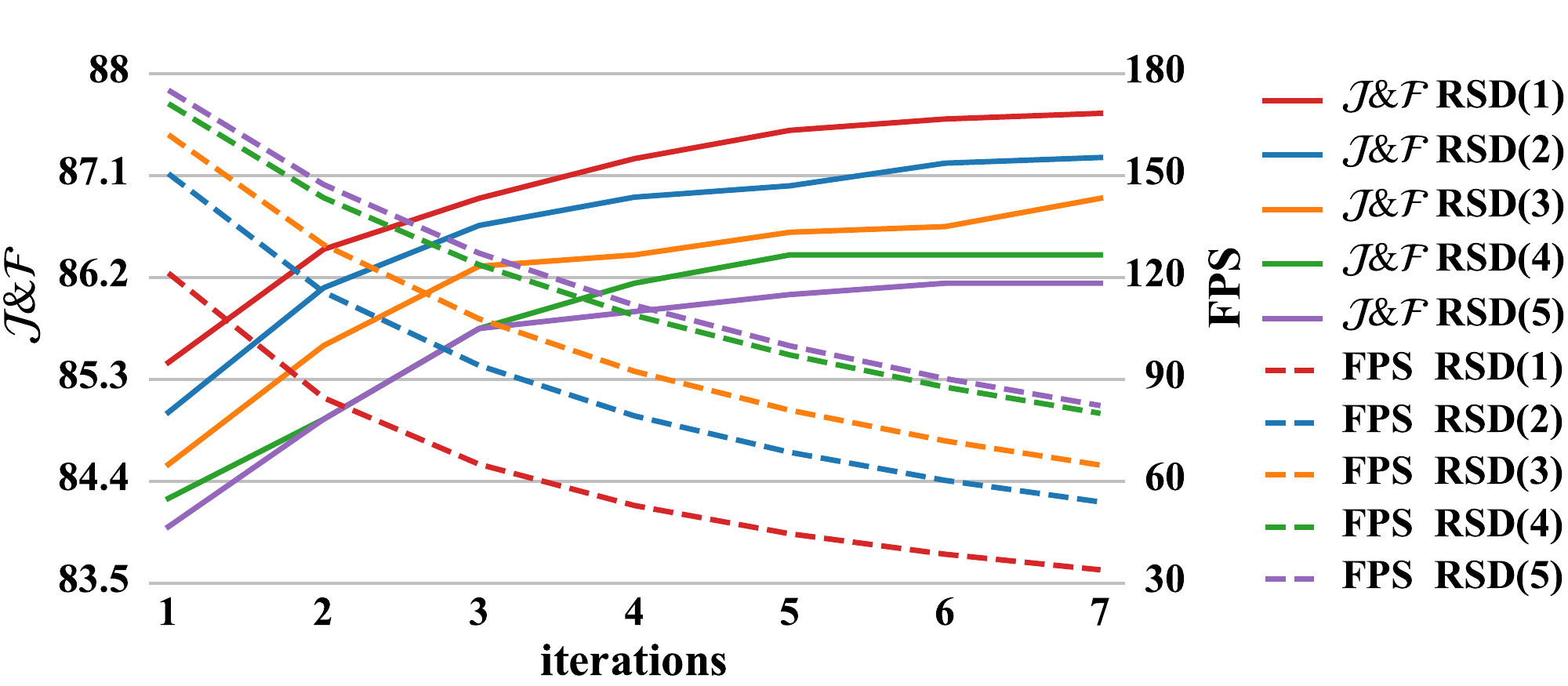}
		\caption{Comparison on DAVIS 2017 validation set}
		\label{fig:rsd-dv17}
	\end{subfigure} %
	
	\caption{Comparison of the optical flow estimated by the Motion-RSD with variation algorithms RSD(1), RSD(2), RSD(3), RSD(4), and RSD(5). $\mathcal{J}$\&$\mathcal{F}$ and FPS are utilized as the evaluation metrics. {\color{black}Iteration} is ranging from 1 to 7.}
\end{figure}

\subsection{Optimization algorithm selection}
\label{sec:selection}
As shown in Figures \ref{fig:rsd-itr} and \ref{fig:flow-update}, $\mathcal{O}_{\omega}$ is close to saturation after around 3 iterations when optimized by the RSD algorithm. Fewer iterations {\color{black}bring} a faster running speed.  To explore a trade-off between accuracy and efficiency for optical flow estimation, we derive a few variations of the original RSD algorithm and compare their performances.

In the original RSD algorithm, optimization is performed every time a new video frame is encountered. This process can be described by Algorithm \ref{alg:A}. The original RSD algorithm can be further accelerated by the parallel processing capability of Pytorch \cite{paszke2017pytorch}. In detail, $\mathcal{O}_{\omega}$ is optimized on multiple training samples each time. Optimization is performed every a fixed interval, not every a frame. This modified procedure can be described in Algorithm \ref{alg:B}. All computations are performed in batch, which is more efficient.

When the update interval $N=1$, Algorithm \ref{alg:B} is equivalent to Algorithm \ref{alg:A}. We refer the variation of RSD with update interval $N$ as RSD($N$). Then, RSD(1), RSD(2), RSD(3), RSD(4), and RSD(5) algorithms are adopted in the Motion-RSD, and their performances {\color{black}in} accuracy and efficiency are compared in {\color{black}Figures} \ref{fig:rsd-dv16} and \ref{fig:rsd-dv17}.

\newcommand{\tabincell}[2]{\begin{tabular}{@{}#1@{}}#2\end{tabular}}
\begin{table*}
\caption{Ablation study of the overall joint learning framework. The learning framework with different {\color{black}conditions} is evaluated on DAVIS 2017 and 2016 validation sets.} 
\label{ablation}
    \centering
    \begin{tabular}{c|cccc|cccc|cccc}
    \toprule
        \multirow{2}{*} {\textbf{No.}} & \multirow{2}{*} {\tabincell{c} {{\textbf{more}} \\ \textbf{data}}} & \multirow{2}{*} {\textbf{frames}} & \multirow{2}{*} {\tabincell{c} {\textbf{motion} \\ \textbf{network}}} & \multirow{2}{*} {\tabincell{c} {\textbf{integration} \\ \textbf{network}}} & \multicolumn{4}{c|} {\bf{DAVIS 2017}} & \multicolumn{4}{c} {\bf{DAVIS 2016}} \\ 
         & & & & & \bm{$\mathcal{J}$}\bf{\&}\bm{$\mathcal{F}$} & \textbf{FPS} & \bm{$\mathcal{J}$} & \bm{$\mathcal{F}$} & \bm{$\mathcal{J}$}\bf{\&}\bm{$\mathcal{F}$} & \textbf{FPS} & \bm{$\mathcal{J}$} & \bm{$\mathcal{F}$}  \\ \midrule

        1 & \xmark & 0 & - & - & 63.9 & 18.43 & 61.4 & 66.4 & 77.5 & 27.62 & 77.5 & 77.4 \\
        2 & \xmark & 1 & RSD* & proposed & 65.6 & 17.02 & 62.8 & 68.3 & 78.2 & 25.91 & 77.8 & 78.5 \\
        3 & \xmark & 1 & SGD-3 & proposed & 64.4 & 15.21 & 61.7 & 67.1 & 77.3 & 23.65 & 77.2 & 77.3 \\
        4 & \xmark & 1 & RAFT-DV & proposed & 66.3 & 5.62 & 63.5 & 69.1 & 78.6 & 6.33 & 78.3 & 78.8 \\
        5 & \xmark & 1 & RSD* & proposed w/o $r$ & 64.9 & 17.11 & 62.0 & 67.8 & 77.0 & 26.05 & 77.0 & 76.9 \\
        6 & \xmark & 2 & RSD* & proposed & 66.4 & 16.74 & 63.3 & 69.5 & 77.8 & 25.51 & 77.3 & 78.3 \\
        7 & \xmark & 2 & RAFT-DV & proposed & 67.5 & 5.59 & 64.3 & 70.7 & 79.0 & 6.29 & 78.5 & 79.4 \\
        9 & \xmark & 3 & RSD* & proposed & 63.4 & 16.46 & 61.0 & 65.7 & 76.8 & 25.13 & 76.4 & 77.2 \\
        10 & \xmark & 3 & RAFT-DV & proposed & 67.5 & 5.56 & 64.4 & 70.6 & 78.8 & 6.26 & 78.2 & 79.4 \\
        12 & \xmark & - & RSD* & GRU & 65.2 & 16.87 & 62.7 & 67.6 & 77.7 & 25.63 & 77.5 & 77.8 \\ \midrule
        13 & \cmark & 0 & - & - & 71.5 & 18.43 & 68.8 & 74.1 & 82.3 & 27.62 & 82.1 & 82.4 \\
        14 & \cmark & 1 & RSD* & proposed & 72.6 & 17.02 & 69.9 & 75.3 & 82.9 & 25.91 & 82.4 & 83.4 \\
        15 & \cmark & 1 & SGD-3 & proposed & 71.7 & 15.21 & 68.9 & 74.4 & 82.0 & 23.65 & 81.7 & 82.3 \\
        16 & \cmark & 1 & RAFT-DV & proposed & 73.2 & 5.62 & 70.5 & 75.8 & 83.2 & 6.33 & 82.6 & 83.8 \\
        17 & \cmark & 1 & RSD* & proposed w/o $r$ & 72.0 & 17.11 & 69.4 & 74.6 & 81.5 & 26.05 & 80.8 & 82.1 \\
        18 & \cmark & 2 & RSD* & proposed & 72.7 & 16.74 & 70.2 & 75.2 & 82.3 & 25.51 & 81.9 & 82.7 \\
        19 & \cmark & 2 & RAFT-DV & proposed & 73.3 & 5.59 & 70.7 & 75.9 & 83.3 & 6.29 & 82.6 & 84.0 \\
        21 & \cmark & 3 & RSD* & proposed & 71.2 & 16.46 & 68.5 & 73.8 & 81.0 & 25.13 & 80.6 & 81.3 \\
        22 & \cmark & 3 & RAFT-DV & proposed & 73.1 & 5.56 & 70.3 & 75.9 & 82.9 & 6.26 & 82.3 & 83.5 \\
        24 & \cmark & - & RSD* & GRU & 72.5 & 16.87 & 69.9 & 75.0 & 82.4 & 25.63 & 82.2 & 82.5 \\ \bottomrule
    \end{tabular}
\end{table*}

As indicated by the dashed lines, the original RSD algorithm (RSD(1)) is improved greatly on FPS by setting an update interval. However, with update interval becoming larger, $\mathcal{J}$\&$\mathcal{F}$ drops dramatically and FPS improvement is getting less. We choose the algorithm RSD(2) with 2 iterations as the balance point, which performs well on $\mathcal{J}$\&$\mathcal{F}$ and FPS. It achieves $\mathcal{J}$\&$\mathcal{F}$ of 86.35 and 86.17 on DAVIS 2016 and 2017 validation sets {\color{black}while} maintaining a very high FPS of 163.03 and 115.72, respectively. Compared {\color{black}with} RSD(1) with 2 or 3 iterations, RSD(2) sacrifices little accuracy (less than 1 on $\mathcal{J}$\&$\mathcal{F}$) but achieves a much faster inference speed (more than 30 FPS). We refer the RSD(2) as RSD*. In the FAMINet, $\mathcal{O}_{\omega}$ of the motion network is optimized by the algorithm RSD*.

\subsection{Ablation study}
\label{sec:ablation}
The ablation study is {\color{black}conducted} on DAVIS 2017 and 2016 datasets. Different experimental conditions are listed on the left of Table \ref{ablation}, and each of them is explained as {\color{black}follows}:
\begin{itemize}
	\item {{\bf{more data}}}: \xmark \ indicates that the FAMINet is trained on DAVIS 2017 training set only. \cmark \ indicates that additional data, Youtube-VOS training set, is utilized for training.
	\item {{\bf{frames}}}: {\color{black}It} indicates {\color{black}the number of previous frames' predictions} utilized for refining the current prediction in the integration network. {\color{black}``0''} indicates that no previous frame is utilized, where the FAMINet degrades to the baseline algorithm, FRTM-fast \cite{robinson2020learning}.
	\item {{\bf{motion network}}}: RSD* indicates that $\mathcal{O}_{\omega}$ in the motion network is optimized by the RSD* as described in Section \ref{sec:selection}. SGD-3 indicates that $\mathcal{O}_{\omega}$ is optimized by the SGD, with learning rate set as 5e-4 and iterations set as 3. RAFT-DV indicates that the motion network is implemented by the RAFT-DV in Section \ref{sec:sanity}.
	\item {{\bf{integration network}}}: {\color{black}``Proposed''} indicates that the integration network is implemented as described in Section \ref{sec:integration}. {\color{black}``Proposed w/o $r$''} indicates that Equations \ref{eq:forget1} and \ref{eq:forget2} are disabled, and Equation \ref{eq:integrate} is modified into $\bar{P}_t = x_t + h_t^{(1)} + h_t^{(2)}$ (if two previous frames' predictions {\color{black}are} considered). {\color{black}``GRU''} indicates that the integration network is implemented by the ConvGRU proposed in \cite{tokmakov2017learning}.
	
\end{itemize}

Results are summarized on the left of Table \ref{ablation}. Comparing No. 1, 2, and 6, we can see that the proposed motion and integration networks improve the baseline FRTM-fast on $\mathcal{J}$\&$\mathcal{F}$ (2.5 percent improvement on DAVIS 2017 validation set) {\color{black}while} maintaining a fast inference speed on par with the baseline. However, when more than 2 previous frames are utilized in the integration network, accuracy is no longer improved and may even be declined. {\color{black}The reason is that} older frames have larger displacements {\color{black}with} respect to the current frame, which is {\color{black}difficult} to be captured by optical flow. Moreover, temporal information of older frames is not as important as adjacent frames for {\color{black}the prediction of the current frame}.

\begin{figure*}[t]
	\centering
	\begin{subfigure}[b]{0.13\linewidth}
	\centering
	\includegraphics[width=24.7mm]{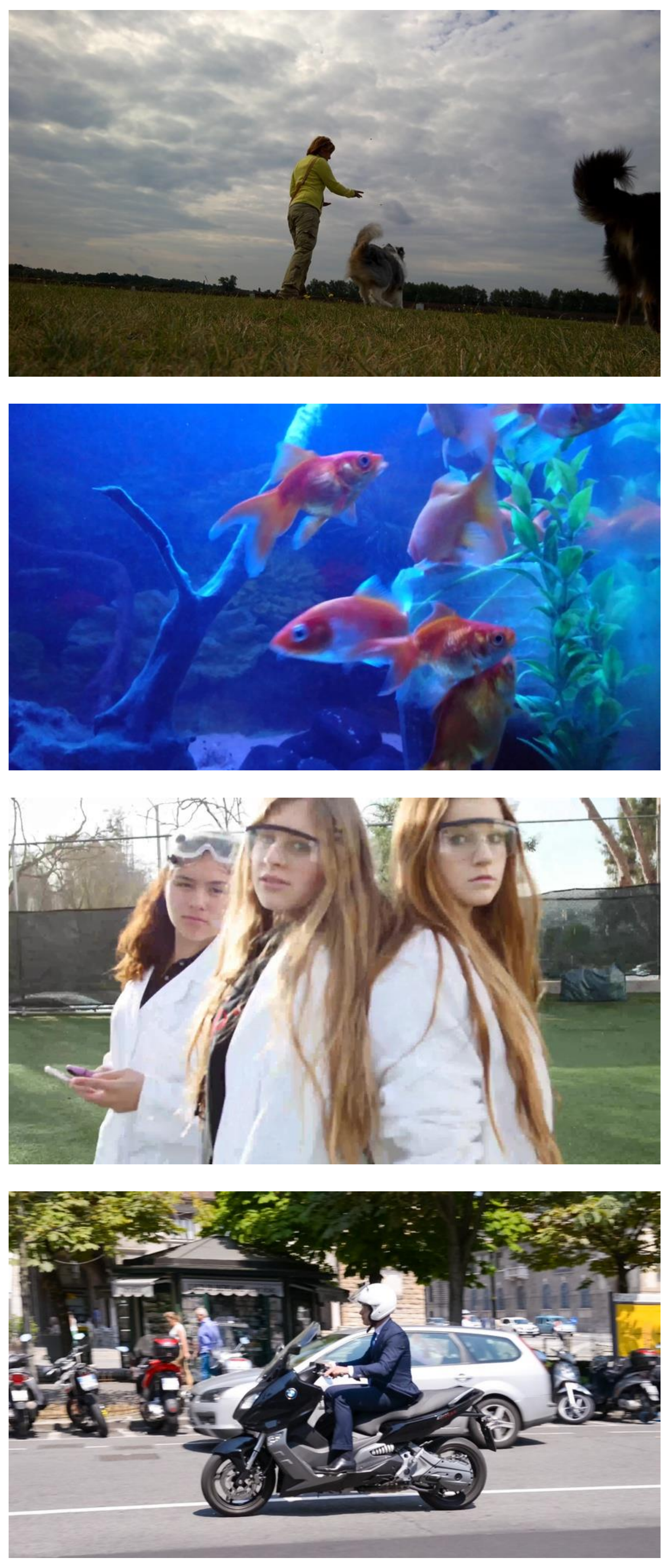}
	\caption{Test image}
	\end{subfigure} %
	\begin{subfigure}[b]{0.13\linewidth}
	\centering
	\includegraphics[width=24.7mm]{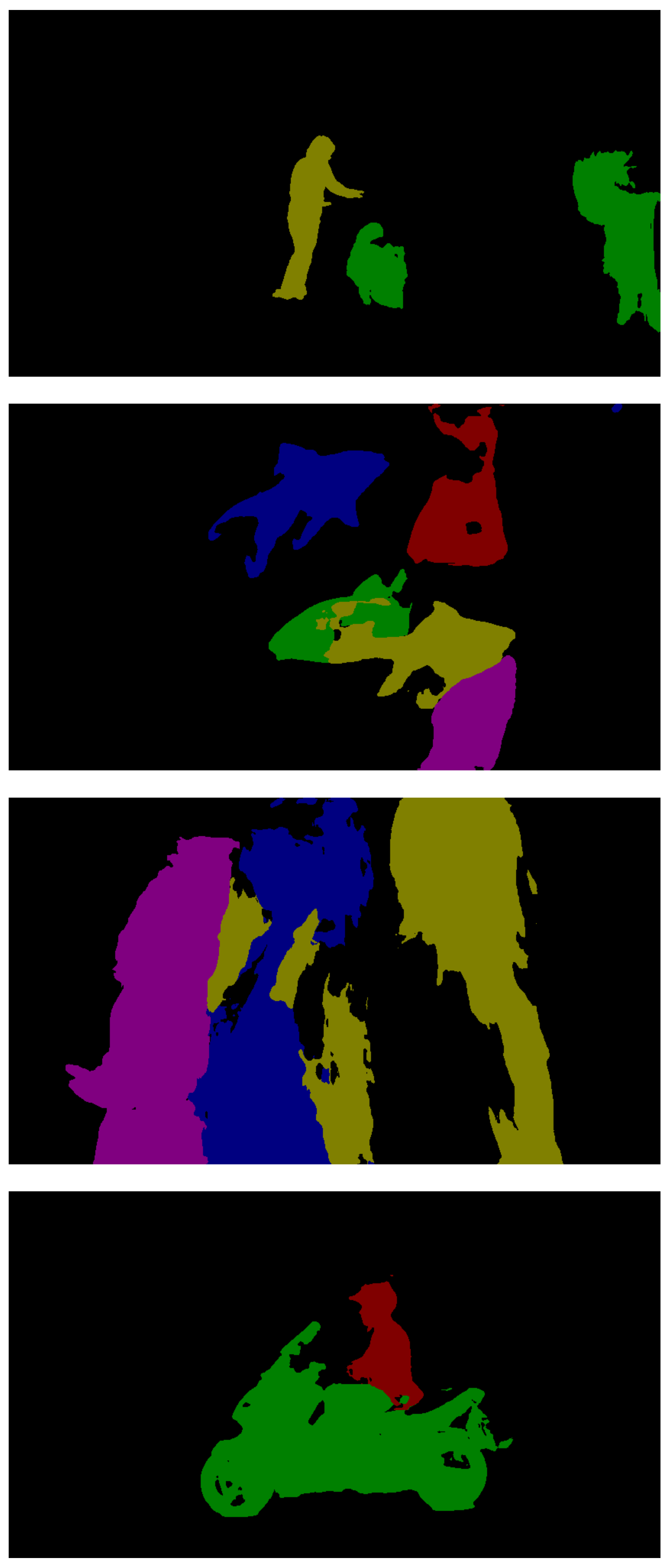}
	\caption{RANet\cite{wang2019ranet}}
	\end{subfigure} %
	\begin{subfigure}[b]{0.13\linewidth}
	\centering
	\includegraphics[width=24.7mm]{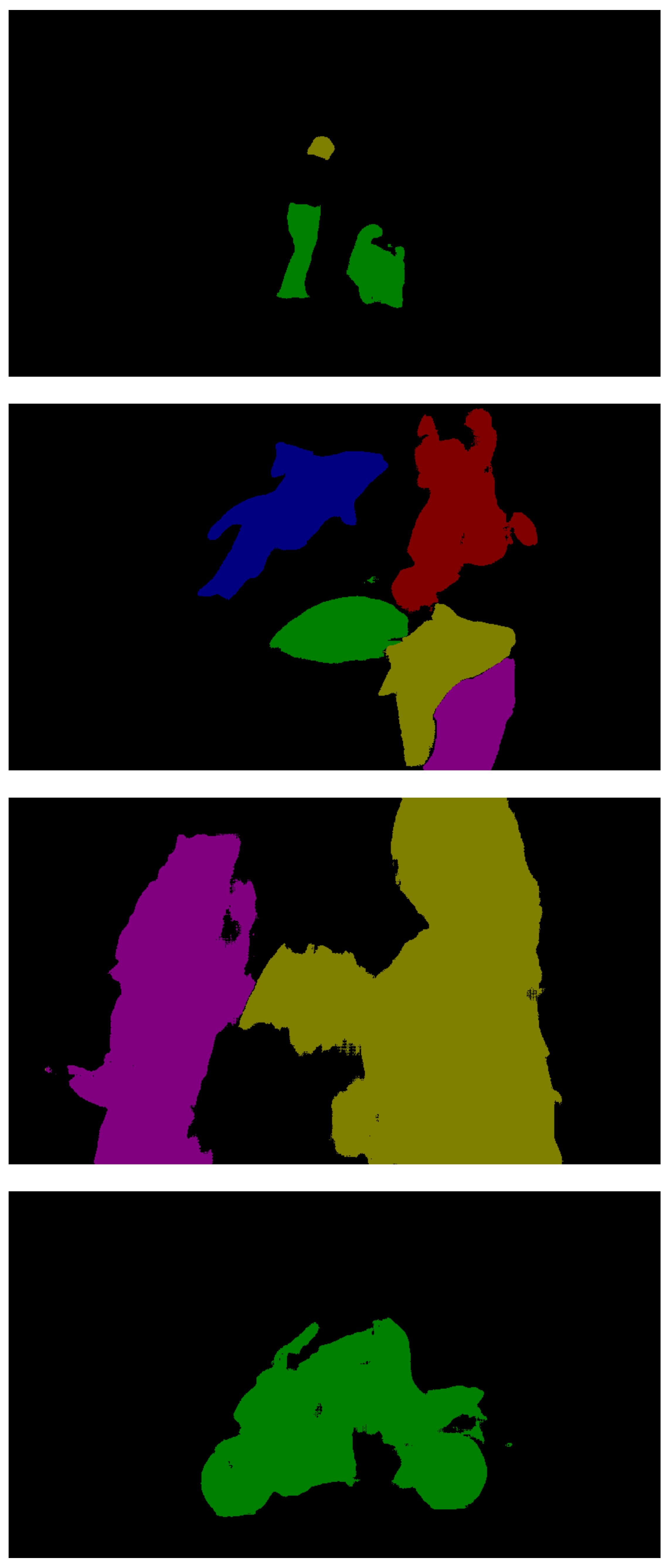}
	\caption{TTVOS\cite{park2020ttvos}}
	\end{subfigure} %
	\begin{subfigure}[b]{0.13\linewidth}
	\centering
	\includegraphics[width=24.7mm]{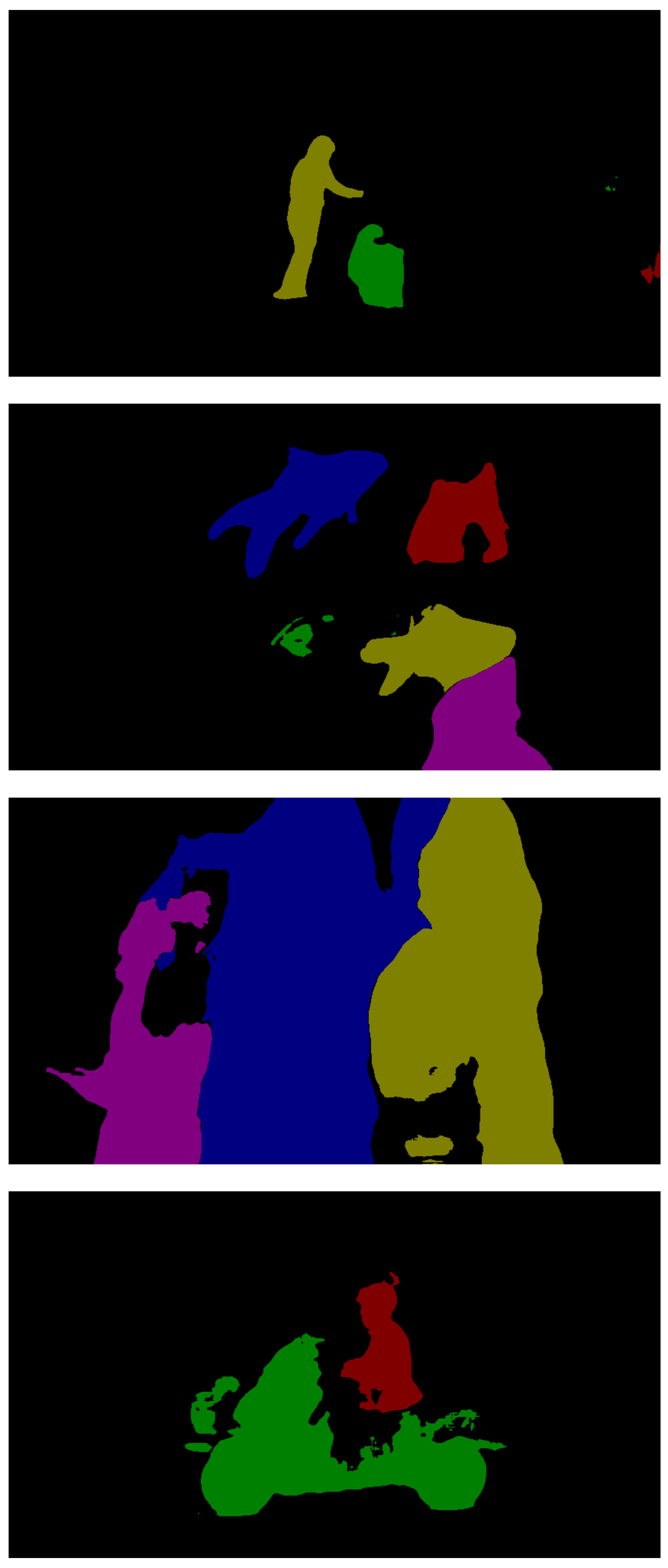}
	\caption{FRTM-fast\cite{robinson2020learning}}
	\end{subfigure} %
	\begin{subfigure}[b]{0.13\linewidth}
	\centering
	\includegraphics[width=24.7mm]{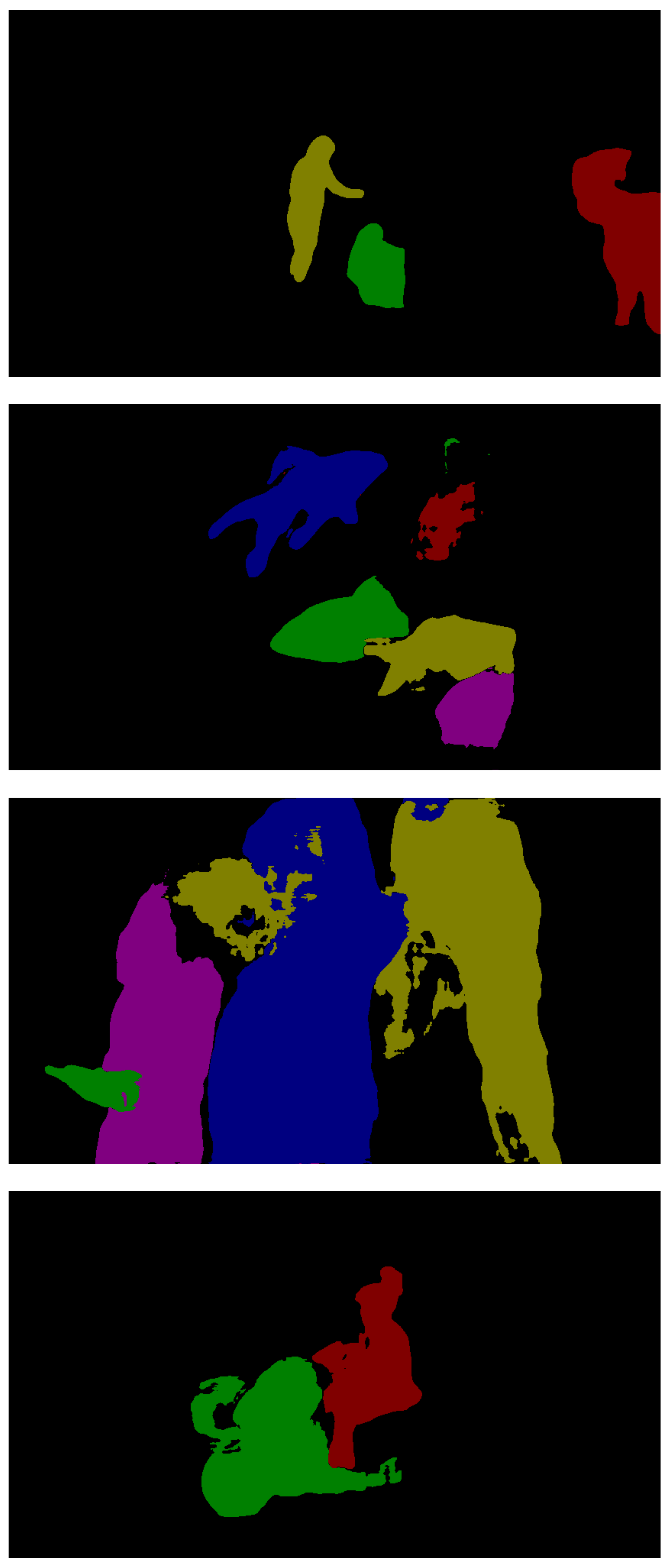}
	\caption{\textcolor{black}{RGF\cite{park2021learning}}}
	\end{subfigure} %
	\begin{subfigure}[b]{0.13\linewidth}
	\centering
	\includegraphics[width=24.7mm]{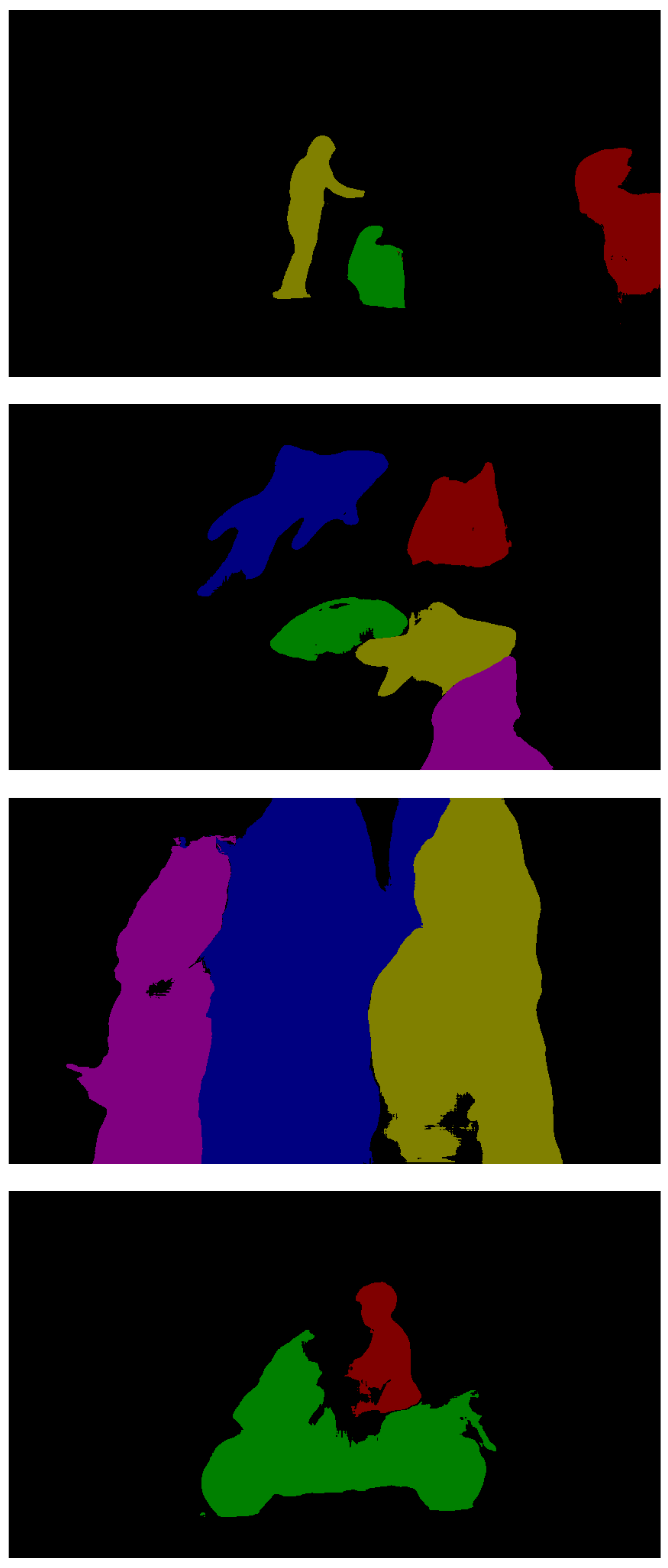}
	\caption{FAMINet-2F}
	\end{subfigure} %
	\begin{subfigure}[b]{0.13\linewidth}
	\centering
	\includegraphics[width=24.7mm]{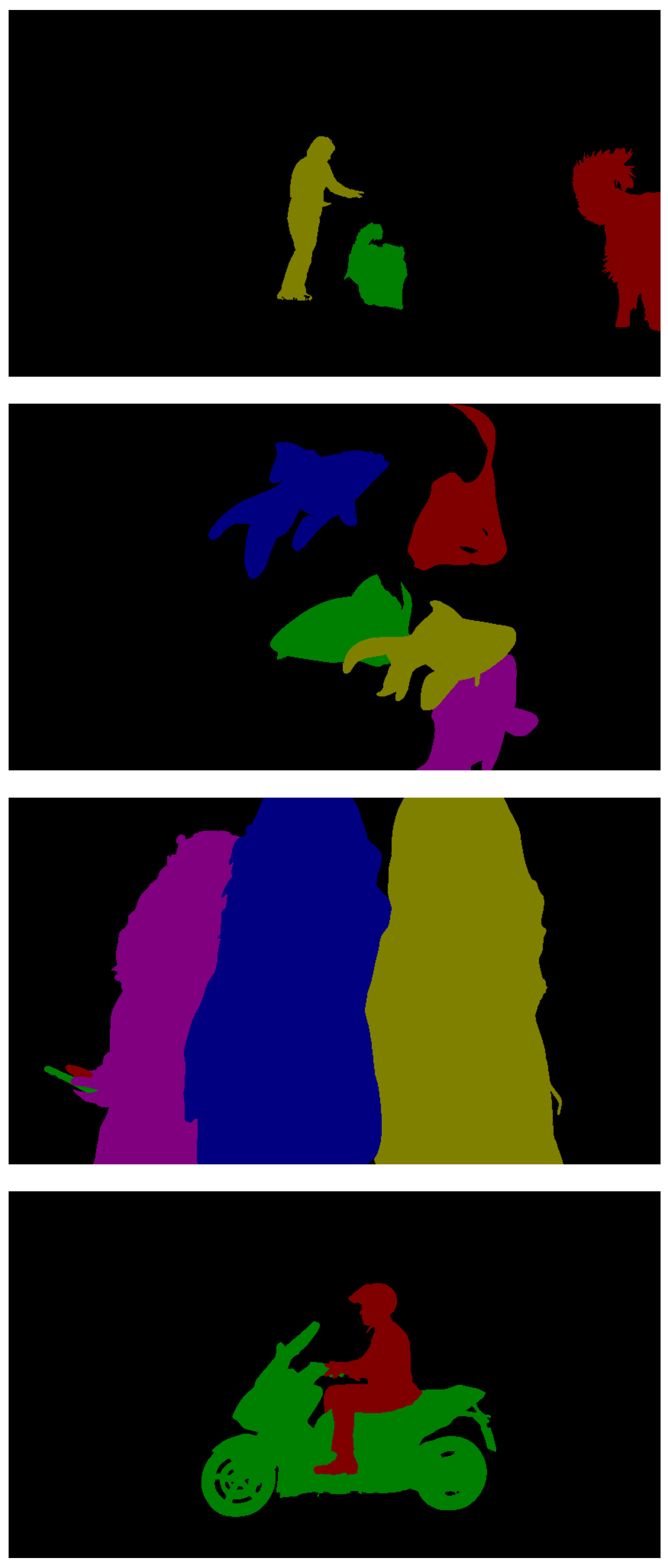}
	\caption{Ground truth}
	\end{subfigure} %
	\caption{\textcolor{black}{Visualization of VOS results by different algorithms. Visualization examples are from DAVIS 2017 validation set. The compared algorithms include additional training data. Different objects are masked in different colors. Ground truth object masks are shown on the right.}}
	\label{fig:vos-dv17}
\end{figure*}

{\bf{Motion network}}. As indicated by No. 3 and 15, introducing {\color{black}information of previous frames} does not bring too many gains when the optical flow is estimated by the SGD-based motion network. {\color{black}The reason is that} the optical flow is of low quality and {\color{black}cannot} register previous predicted object probability map to the current frame accurately. On the contrary, No. 7 and 19 achieve the best results on $\mathcal{J}$\&$\mathcal{F}$ as the RAFT-DV produces the most accurate optical flow. Compared to the RAFT-DV, the proposed RSD* for the FAMINet sacrifices little on accuracy but leads {\color{black}to three to four} times faster inference speed.

{\bf{Integration network}}. As indicated by No. 5 and 17, $\mathcal{J}$\&$\mathcal{F}$ decreases when $r_t$ is abandoned. On DAVIS 2016 validation set, performances of No. 5 and 17 are even worse than the baseline. Therefore, $r_t$ plays an important role in the integration network. It works as a confidence map, which can suppress temporal information propagated by inaccurate optical flow. Comparing No. 2 and 12, or comparing No. 14 and 24, we can see that $\mathcal{J}$\&$\mathcal{F}$ decreases a little when the integration network is replaced by the ConvGRU. However, the ConvGRU needs to be trained on relatively long sequences \cite{tokmakov2017learning}, which takes up a lot of GPU memories. In addition, the proposed integration network includes mask propagation and confidence computation, which is more explainable.

Compared {\color{black}with} DAVIS 2016 validation set, the FAMINet improves more from the baseline FRTM-fast on DAVIS 2017 validation set. {\color{black}The reason is that} DAVIS 2017 includes multi-object tracking and is more challenging. The cases of occlusion, large displacement, and object deformation are very common. Thus, the proposed motion and integration networks are more in need of solving these problems.

\subsection{{\color{black}Comparison with state-of-the-art methods}}
The FAMINet is compared with other state-of-the-art VOS methods. They are evaluated on the DAVIS and YouTubeVOS benchmarks. For the FAMINet utilizing $n$ previous frames, we denote it as FAMINet-$n$F. We choose FAMINet-1F and FAMINet-2F for comparison {\color{black}given that} they perform well on accuracy and efficiency.

\subsubsection{DAVIS}
For the DAVIS benchmark, VOS methods are first evaluated on DAVIS 2016 and 2017 validation sets, as shown in Table \ref{dvvalidation}. {\color{black}Notably,} a few VOS methods {\color{black}require} additional training data or pre-training on segmentation data. For a fair comparison, we classify methods into two categories: methods trained on DAVIS 2017 training set only, and methods trained on DAVIS and additional data. As shown on the middle of Table \ref{dvvalidation}, {\bf{yt}} indicates YoutubeVOS training set, {\bf{seg}} indicates the segmentation datasets (Pascal or COCO), and {\bf{syn}} indicates a saliency dataset for making synthetic video clip.

For comparing the efficiency, the FPS is utilized as the metric. Following previous methods, the FPS is measured on DAVIS 2016 validation set. However, the inference speed of algorithms varies divergently on different hardwares. For a fair comparison, a few VOS methods are re-implemented on our environment (NVIDIA TITAN RTX GPU). Re-implemented methods are marked with $\diamond$. We also report their original results on the {\color{black}bottom part} of Table \ref{dvvalidation} for reference. The results of other methods are directly quoted from their original papers.

\begin{table}
\caption{Comparisons with state-of-the-art VOS methods on DAVIS 2017 and 2016 validation sets (abbreviated as DV17 and DV16, respectively). Our proposed algorithm is shown in {\bf{bold}}. The best and second best are shown in {\bf{bold}} and \underline{underlined}.}
\label{dvvalidation}
    \centering
    \begin{tabular}{l|ccc|cc|c}
    \toprule
        \multirow{2}{*} {\textbf{method}} & \multicolumn{3}{c|} {\bf{more data}} & \multicolumn{1}{c}{\bf{DV17}} & \multicolumn{1}{c|}{\bf{DV16}} & \multicolumn{1}{c}{\bf{DV16}} \\
        & \bf{yv} & \bf{seg} & \bf{syn} & \bm{$\mathcal{J}$}\bf{\&}\bm{$\mathcal{F}$} & \bm{$\mathcal{J}$}\bf{\&}\bm{$\mathcal{F}$} & \bf{FPS} \\ \midrule
        STM\cite{oh2019video} & \xmark & \xmark & \xmark & 43.0 & - & 6.25 \\
        FAVOS\cite{cheng2018fast} & \xmark & \xmark & \xmark & 58.2 & {\bf{80.8}} & 0.56 \\
        AGAME\cite{johnander2019generative} & \xmark & \xmark & \xmark & 63.2 & - & 14.30 \\
        FRTM-fast$\diamond$\cite{robinson2020learning} & \xmark & \xmark & \xmark & \textcolor{black}{63.9} & 77.5 & \underline{22.11} \\
         \textcolor{black}{SiamDyna\cite{hong2021siamese}} & \xmark & \xmark & \xmark & 61.4 & 72.3 & {\bf{35.00}} \\
        \bf{FAMINet-1F} & \xmark & \xmark & \xmark & \underline{65.6} & \underline{78.2} & \textcolor{black}{20.58} \\
        \bf{FAMINet-2F} & \xmark & \xmark & \xmark & {\bf{66.4}} & \textcolor{black}{77.8} & 20.25 \\ \midrule
        RVOS\cite{ventura2019rvos} & \cmark & \xmark & \xmark & 50.3 & - & 22.70 \\
        OSMN\cite{yang2018efficient} & \xmark & \cmark & \xmark & 54.8 & 73.5 & 7.69 \\
        RANet$\diamond$\cite{wang2019ranet} & \xmark & \xmark & \cmark & 65.7 & 85.5 & 10.87 \\
        RGMP\cite{oh2018fast} & \xmark & \xmark & \cmark & 66.7 & 81.8 & 7.70 \\
        DTN\cite{zhang2019fast} & \xmark & \cmark & \xmark & 67.4 & 83.6 & 14.30 \\
        TTVOS$\diamond$\cite{park2020ttvos} & \cmark & \xmark & \cmark & 67.8 & 83.8 & \underline{26.84} \\
        OnAVOS\cite{voigtlaender2017online} & \xmark & \cmark & \xmark & 67.9 & 85.5 & 0.08 \\
        OSVOS-S\cite{maninis2018video} & \xmark & \cmark & \xmark & 68.0 & \underline{86.5} & 0.22 \\
        AGAME\cite{johnander2019generative} & \cmark & \xmark & \cmark & 70.0 & 82.1 & 14.30 \\
        FEELVOS\cite{voigtlaender2019feelvos} & \cmark & \cmark & \xmark & 71.5 & 81.7 & 2.22 \\
        MGCRN\cite{hu2020motion} & \xmark & \cmark & \xmark & - & 85.1 & 1.37 \\
        STM\cite{oh2019video} & \cmark & \xmark & \cmark & {\bf{81.8}} & {\bf{89.3}} & 6.25 \\
        FRTM-fast$\diamond$\cite{robinson2020learning} & \cmark & \xmark & \xmark & 71.5 & 82.3 & {\bf{27.62}} \\
        \textcolor{black}{TVOS\cite{zhang2020transductive}} & \cmark & \xmark & \xmark & 72.3 & - & -  \\
        RGF$\diamond$\cite{park2021learning} & \cmark & \xmark & \xmark & 72.1 & 80.8 & 25.15 \\
        \textcolor{black}{AGUNet\cite{yin2021agunet}} & \xmark & \cmark & \xmark & 64.1 & 80.9 & 11.11  \\
        \textcolor{black}{LLGC\cite{liang2021semi}} & \cmark & \xmark & \xmark & 69.5 & - & -  \\
        \textcolor{black}{OGS\cite{fan2021semi}} & \cmark & \xmark & \xmark & 71.8 & 82.3 & -  \\
        \textcolor{black}{STG-Net\cite{liu2021spatiotemporal}} & \xmark & \cmark & \xmark & \underline{74.7} & \textcolor{black}{85.7} & 6.00  \\
        \bf{FAMINet-1F} & \cmark & \xmark & \xmark & 72.6 & 82.9 & \textcolor{black}{25.91} \\
        \bf{FAMINet-2F} & \cmark & \xmark & \xmark & \textcolor{black}{72.7} & 82.3 & 25.51 \\ \midrule
        RANet\cite{wang2019ranet} & \xmark & \xmark & \cmark & 65.7 & 85.5 & 30.30 \\
        TTVOS\cite{park2020ttvos} & \cmark & \xmark & \cmark & 67.8 & 83.8 & 39.60 \\
        FRTM-fast\cite{robinson2020learning} & \cmark & \xmark & \xmark & 70.2 & 78.5 & 41.30 \\
        RGF\cite{park2021learning} & \cmark & \xmark & \xmark & 71.7 & 80.9 & 37.60 \\ \bottomrule
    \end{tabular}
\end{table}

Compared with state-of-the-art methods, our proposed method achieves comparable or even better performances on $\mathcal{J}$\&$\mathcal{F}$ but via a much faster inference speed, with or without additional data. STM \cite{oh2019video} performs the best on accuracy, but it is {\color{black}time consuming}. Compared {\color{black}with} the most efficient methods TTVOS \cite{park2020ttvos} and FRTM-fast \cite{robinson2020learning}, our method outperforms them on $\mathcal{J}$\&$\mathcal{F}$ with a comparable inference speed. Some visualization results are shown in Figure \ref{fig:vos-dv17}. We can see that our method has great advantages in multi-object segmentation. Temporal information is exploited against the missing or incomplete segmentations for objects (e.g., the missing segmentations for dog on the first row, fish on the second row, and the incomplete segmentations for people on the third row, the motorbike on the fourth row). Benefitting from the proposed motion and integration networks, the FAMINet produces object segmentations with better consistency.

We conduct evaluations on DAVIS 2017 test split, as shown in Table \ref{dvtest}. {\bf{OL}} indicates that online learning for VOS is included in the algorithm during test. {\color{black}Notably,} online learning for appearances and motions of objects are included in the FAMINet. Our methods achieve the second and third best results compared with state-of-the-art VOS methods.

\textcolor{black}{We compare the FAMINet with state-of-the-art optical flow-based VOS methods, as shown in Table \ref{dv16flow}. The optical flow methods adopted by these methods are also reported. We can see that most of these VOS methods have a very low FPS. For the VOS methods that adopt traditional optical flow methods, they can spend several seconds processing a video frame, which is inapplicable to real-time systems. Meanwhile our method reaches comparable accuracy in real time. Although the SegFlow\cite{cheng2017segflow} reaches an equivalent running speed to ours, it performs worse on accuracy (76.1 on $\mathcal{J}$\&$\mathcal{F}$)}.

\begin{table}
\caption{Comparisons with state-of-the-art VOS methods on DAVIS 2017 test split. Our proposed algorithm is shown in {\bf{bold}}. The best and second best for $\mathcal{J}$\&$\mathcal{F}$ are shown in {\bf{bold}} and \underline{underlined}.}
\label{dvtest}
    \centering
    \begin{tabular}{l|c|ccc}
    \toprule
        \bf{method} & \bf{OL} & \bm{$\mathcal{J}$}\bf{\&}\bm{$\mathcal{F}$} & \bm{$\mathcal{J}$} & \bm{$\mathcal{F}$} \\ \midrule
        OSMN\cite{yang2018efficient} & \xmark & 41.3 & 37.7 & 44.9 \\
        SiamMask\cite{wang2019fast} & \xmark & 43.2 & 40.6 & 45.8 \\
        FAVOS\cite{cheng2018fast} & \xmark & 43.6 & 42.9 & 44.2 \\
        RVOS\cite{ventura2019rvos} & \xmark & 50.3 & 48.0 & 52.6 \\
        AGAME\cite{johnander2019generative} & \xmark & 52.3 & 49.2 & 55.3 \\
        RGMP\cite{oh2018fast} & \xmark & 52.9 & 51.3 & 54.4 \\
        RANet\cite{wang2019ranet} & \xmark & 55.3 & 53.4 & 57.2 \\
        AGSS\cite{lin2019agss} & \xmark & 57.2 & 54.8 & 59.7 \\
        FEELVOS\cite{voigtlaender2019feelvos} & \xmark & 57.8 & 55.1 & 60.4 \\
        OSVOS\cite{caelles2017one} & \cmark & 50.9 & 47.0 & 54.8 \\
        OnAVOS\cite{voigtlaender2017online} & \cmark & 52.8 & 49.9 & 55.7 \\
        OSVOS-S\cite{maninis2018video} & \cmark & 57.5 & 52.9 & 62.1 \\
        FRTM-fast$\diamond$\cite{robinson2020learning} & \cmark & 58.6 & 56.1 & 61.2 \\
        \textcolor{black}{STG-Net\cite{liu2021spatiotemporal}} & \xmark & {\bf{63.1}} & 59.7 & 66.5 \\
        \bf{FAMINet-1F} & \cmark & \underline{59.7} & 57.0 & 62.3 \\
        \bf{FAMINet-2F} & \cmark & \textcolor{black}{59.6} & 56.6 & 62.7 \\ \bottomrule
    \end{tabular}
\end{table}

\begin{table}
\caption{ \textcolor{black}{Comparisons with state-of-the-art optical flow-based VOS methods on DAVIS 2016 validation set. Our proposed algorithm is {\bf{bolded}}. The best and second best for $\mathcal{J}$\&$\mathcal{F}$ and FPS are shown in {\bf{bold}} and \underline{underlined}.}}
\label{dv16flow}
    \centering
    \begin{tabular}{l|l|cccc}
    \toprule
        \bf{method} & \bf{Optical flow} & \bm{$\mathcal{J}$}\bf{\&}\bm{$\mathcal{F}$} & \bf{FPS} & \bm{$\mathcal{J}$} & \bm{$\mathcal{F}$} \\ \midrule
        MP-Net\cite{tokmakov2017learning2} & LDOF\cite{brox2010large} & 68.0 & 0.02 & 69.7 & 66.3 \\
        ObjFlow \cite{tsai2016video} & ClassicNL\cite{sun2014quantitative}& 69.5 & 0.01 & 71.1 & 67.9 \\
        Fusionseg\cite{jain2017fusionseg} & Beyond\cite{liu2009beyond}& -  & 0.17 & 71.5 & - \\
        LVO\cite{tokmakov2017learning} & LDOF\cite{brox2010large}& 74.1 & 0.02 & 75.9 & 72.1 \\
        MaskTrack\cite{perazzi2017learning} & EpicFlow\cite{revaud2015epicflow}& - & 0.04 & 78.4 & - \\
        SegFlow\cite{cheng2017segflow} & FlowNetS\cite{dosovitskiy2015flownet}& 76.1 & \underline{25.51} & 76.1 & 76.0 \\
        Lucid\cite{khoreva2019lucid} & FlowNet2\cite{ilg2017flownet}& \underline{85.7} & 0.41 & 86.6 & 84.8 \\
        PReMVOS\cite{luiten2018premvos} & FlowNet2\cite{ilg2017flownet}& {\bf{86.8}} & 1.71 & 84.9 & 88.6 \\
        MG\cite{hu2020motion} & FlowNet2\cite{ilg2017flownet}& \textcolor{black}{85.1} & \textcolor{black}{6.74} & 84.4 & 85.7 \\
        {\bf{FAMINet-1F}} & \bm{$\mathcal{O}_{\omega}$}{\bf{-RSD*}}& 82.9 & {\bf{25.91}} & 82.4 & 83.4 \\ \bottomrule

    \end{tabular}
\end{table}

\begin{table}
\caption{Comparisons with state-of-the-art VOS methods on Youtube-VOS validation. Our proposed algorithm is shown in {\bf{bold}}. The best and second best for $\mathcal{J}$\&$\mathcal{F}$ are shown in {\bf{bold}} and \underline{underlined}.}
\label{ytvalidation}
    \centering
    \begin{tabular}{l|cc|c|cc|cc}
    \toprule
        \multirow{2}{*} {\textbf{method}} & \multicolumn{2}{c|} {\bf{more data}} & \bm{$\mathcal{J}$}\bf{\&}\bm{$\mathcal{F}$} & \multicolumn{2}{c|}{\bm{$\mathcal{J}$}} & \multicolumn{2}{c}{\bm{$\mathcal{F}$}} \\
        & \bf{seg} & \bf{syn} & \multicolumn{-1}{c|}{ \bf{all}} & \bf{S} & \bf{unS} & \bf{S} & \bf{unS} \\ \midrule
        OnAVOS\cite{voigtlaender2017online} & \cmark & \xmark & 55.2 & 60.1 & 46.1 & 62.7 & 51.4 \\
        OSVOS\cite{caelles2017one} & \cmark & \xmark & 58.8 & 59.8 & 54.2 & 60.5 & 60.7 \\
        RVOS\cite{ventura2019rvos} & \xmark & \xmark & 56.8 & 63.6 & 45.5 & 67.2 & 51.0 \\
        S2S\cite{xu2018youtube} & \xmark & \xmark & 64.4 & 71.0 & 55.5 & 70.0 & 61.2 \\
        AGAME\cite{johnander2019generative} & \xmark & \cmark & 66.1 & 67.8 & 60.8 & 69.5 & 66.2 \\
        STM\cite{oh2019video} & \xmark & \cmark & {\bf{79.4}} & 79.7 & 72.8 & 84.2 & 80.9 \\
        PReMVOS\cite{luiten2018premvos} & \cmark & \cmark & 66.9 & 71.4 & 56.5 & - & - \\
        FRTM-fast\cite{robinson2020learning} & \xmark & \xmark & 65.7 & 68.6 & 58.4 & 71.3 & 64.6 \\
        FRTM-fast$\diamond$ & \xmark & \xmark & 64.9 & 68.1 & 57.2 & 70.8 & 63.5 \\
        RGF\cite{park2021learning} & \xmark & \cmark & 63.8 & 68.3 & 55.2 & 70.6 & 61.0 \\
        \textcolor{black}{LLGC\cite{liang2021semi}} & \xmark & \xmark & 66.4 & 66.2 & 68.6 & 61.4 & 69.2 \\
        \textcolor{black}{OGS\cite{fan2021semi}} & \xmark & \xmark & \textcolor{black}{67.2} & 67.0 & 62.1 & 69.6 & 70.2 \\
        \textcolor{black}{STG-Net\cite{liu2021spatiotemporal}} & \cmark & \xmark & \underline{73.0} & 72.7 & 69.1 & 75.2 & 74.9 \\
        \bf{FAMINet-1F} & \xmark & \xmark & 66.2 & 69.1 & 58.8 & 71.5 & 65.4 \\
        \bf{FAMINet-2F} & \xmark & \xmark & 67.1 & 69.2 & 60.5 & 71.5 & 67.2 \\ \bottomrule
    \end{tabular}
\end{table}

\begin{figure*}[t]
	\centering
	\begin{subfigure}[b]{0.13\linewidth}
	\centering
	\includegraphics[width=24.7mm]{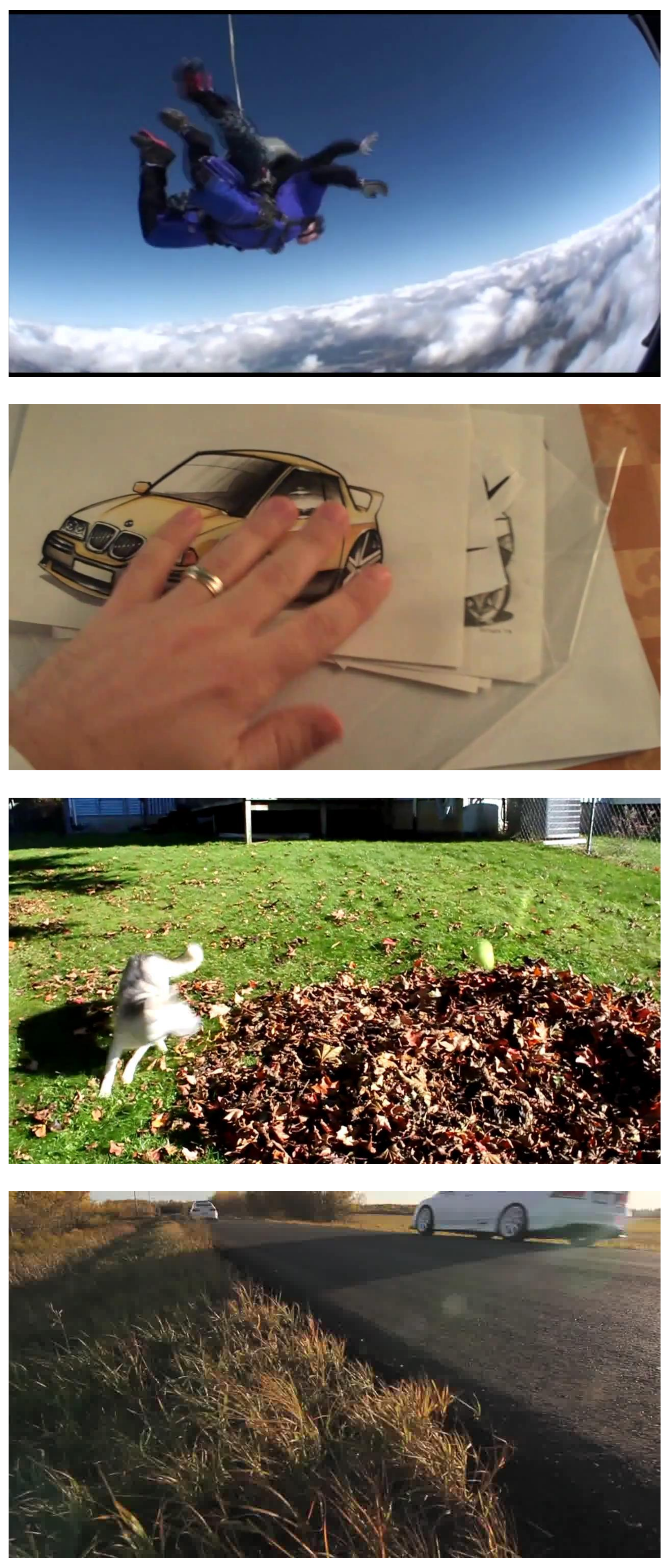}
	\caption{Test image}
	\end{subfigure} %
	\begin{subfigure}[b]{0.13\linewidth}
	\centering
	\includegraphics[width=24.7mm]{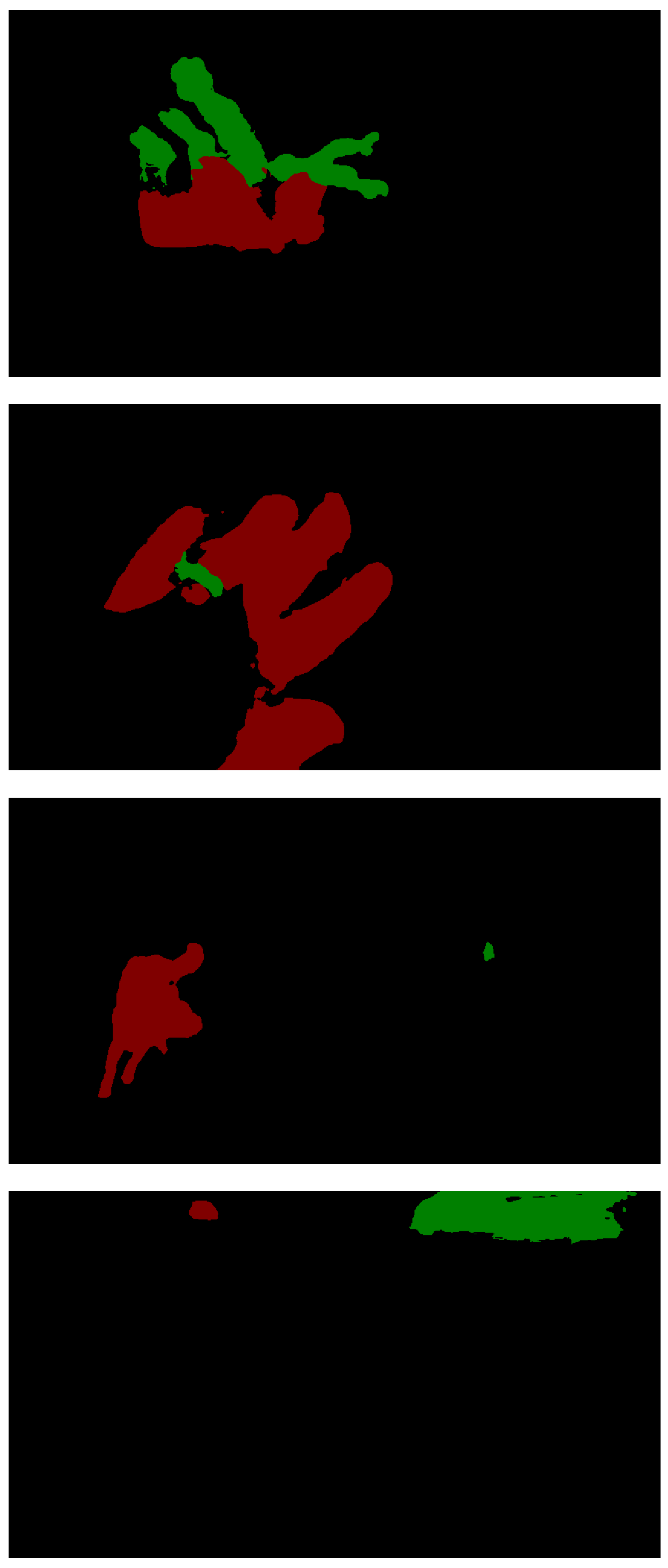}
	\caption{RANet\cite{wang2019ranet}}
	\end{subfigure} %
	\begin{subfigure}[b]{0.13\linewidth}
	\centering
	\includegraphics[width=24.7mm]{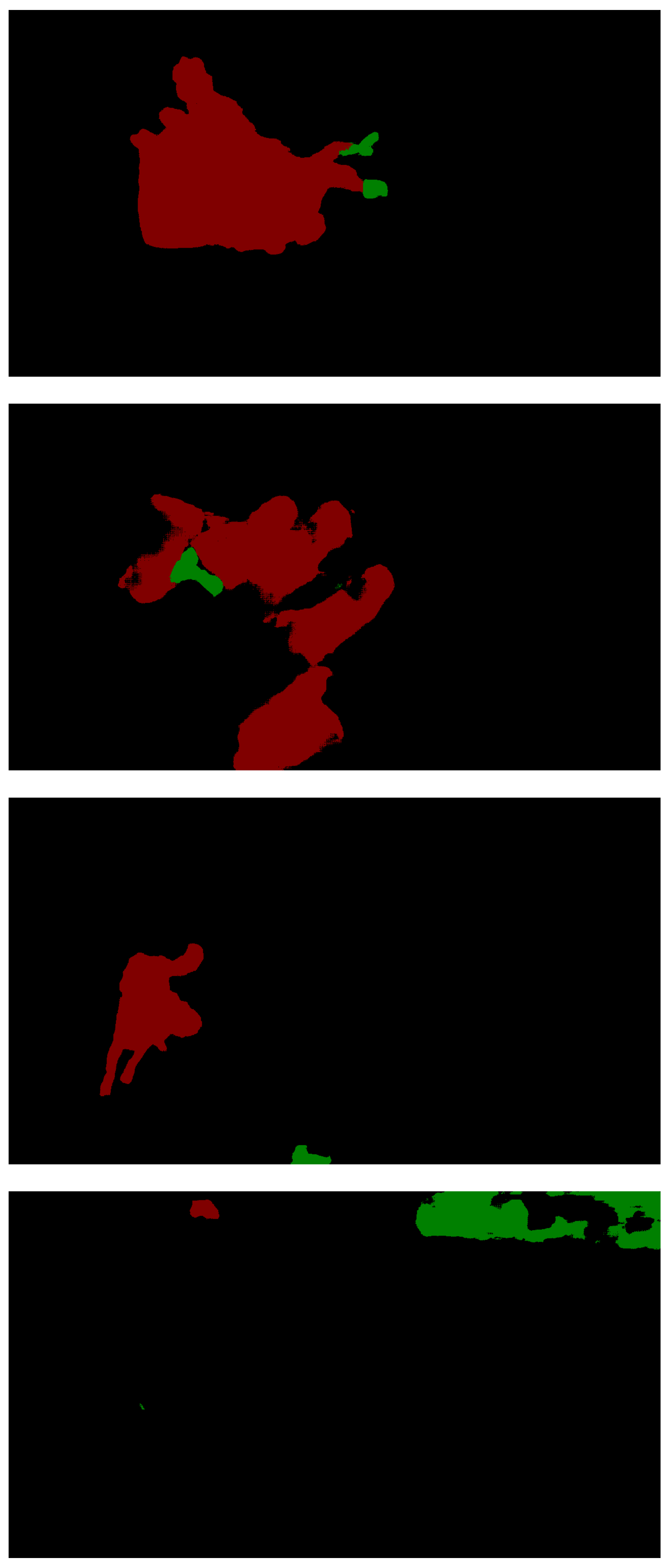}
	\caption{TTVOS\cite{park2020ttvos}}
	\end{subfigure} %
	\begin{subfigure}[b]{0.13\linewidth}
	\centering
	\includegraphics[width=24.7mm]{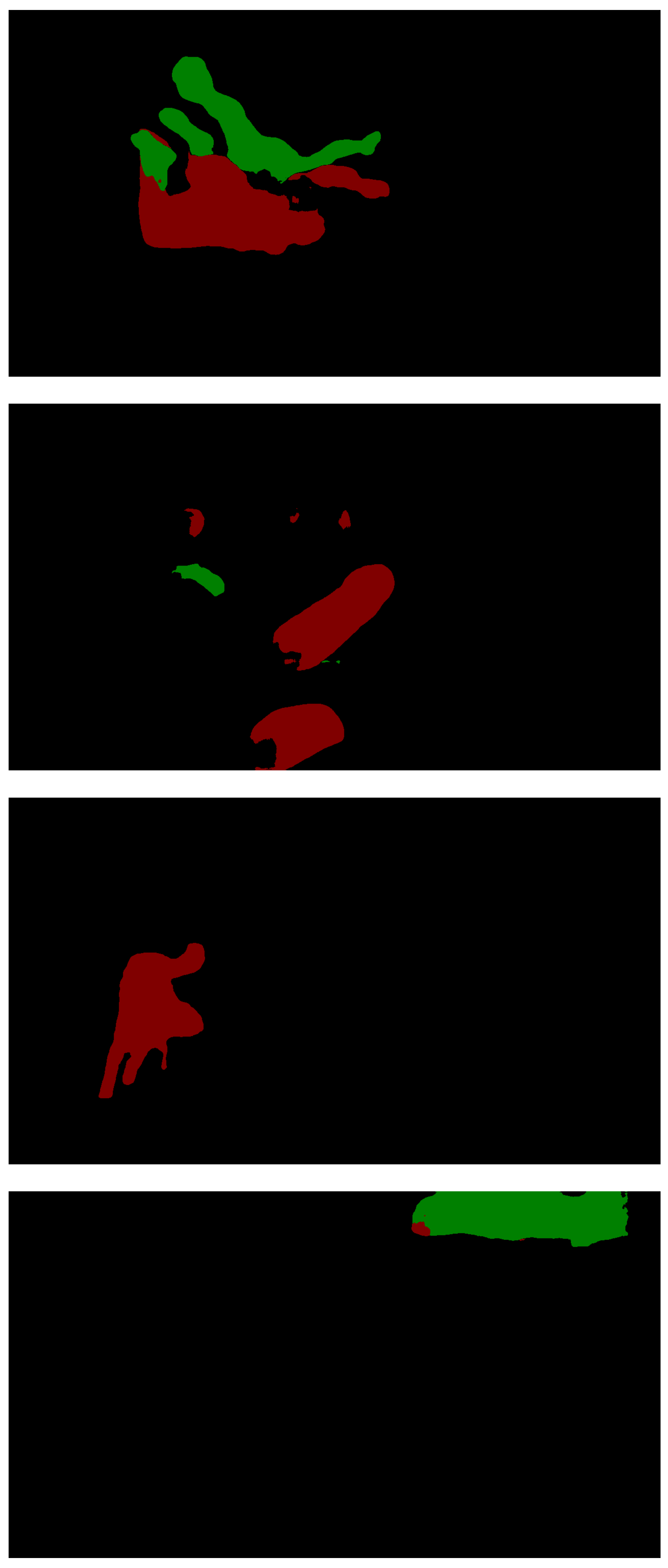}
	\caption{FRTM-fast\cite{robinson2020learning}}
	\end{subfigure} %
	\begin{subfigure}[b]{0.13\linewidth}
	\centering
	\includegraphics[width=24.7mm]{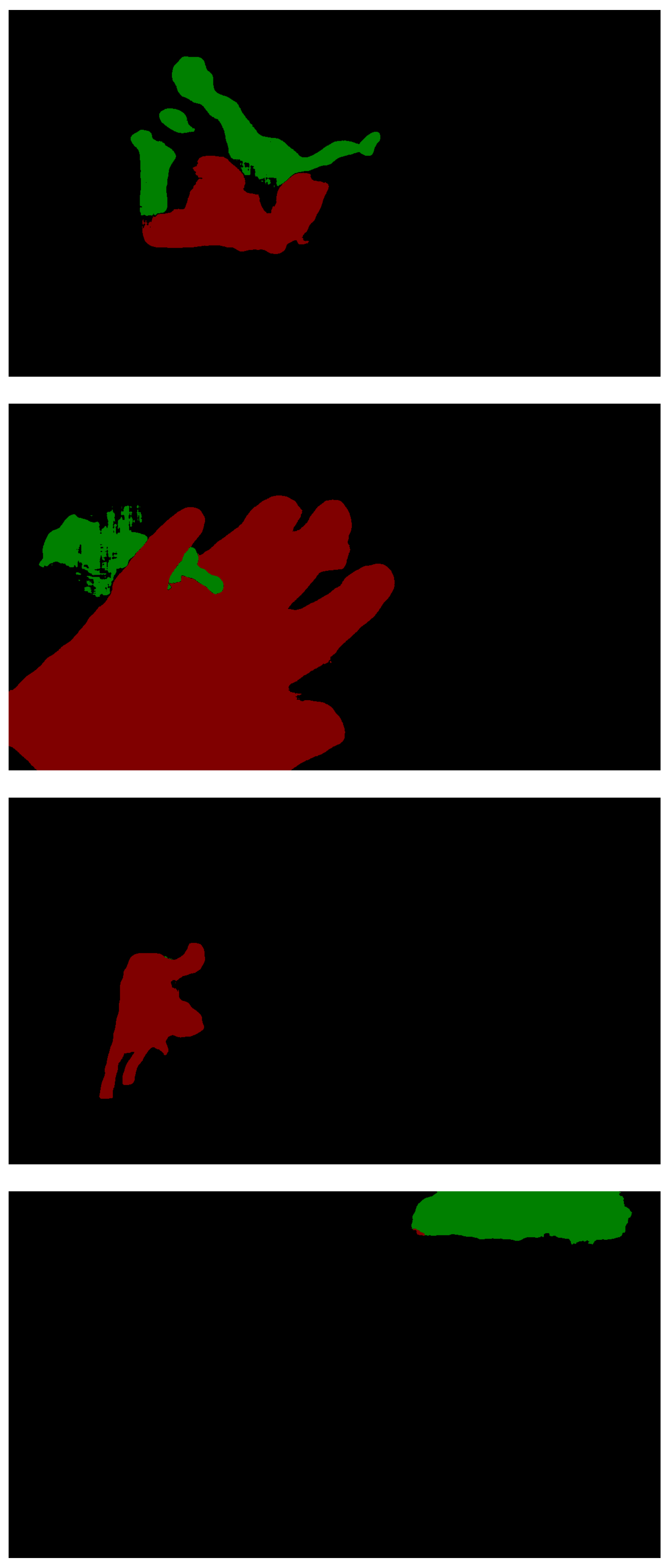}
	\caption{\textcolor{black}{RGF\cite{park2021learning}}}
	\end{subfigure} %
	\begin{subfigure}[b]{0.13\linewidth}
	\centering
	\includegraphics[width=24.7mm]{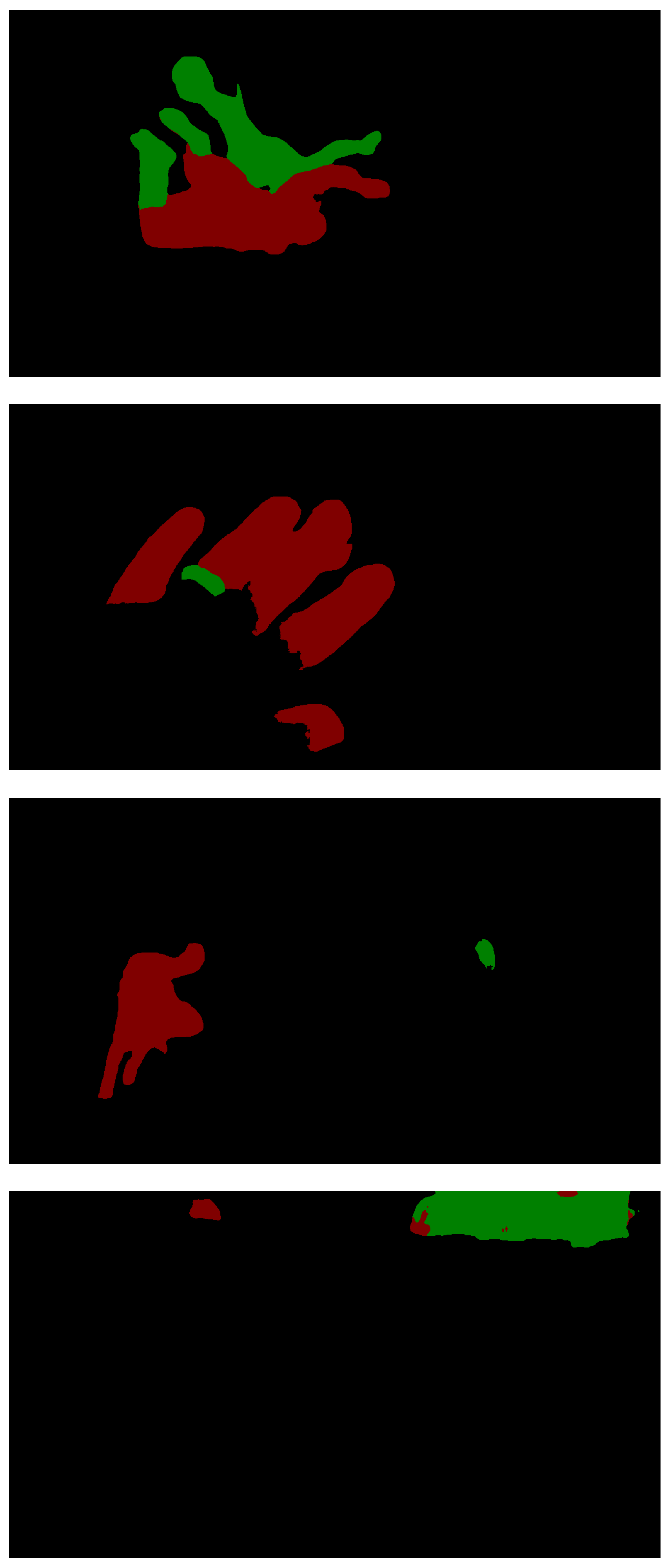}
	\caption{FAMINet-2F}
	\end{subfigure} %
	\begin{subfigure}[b]{0.13\linewidth}
	\centering
	\includegraphics[width=24.7mm]{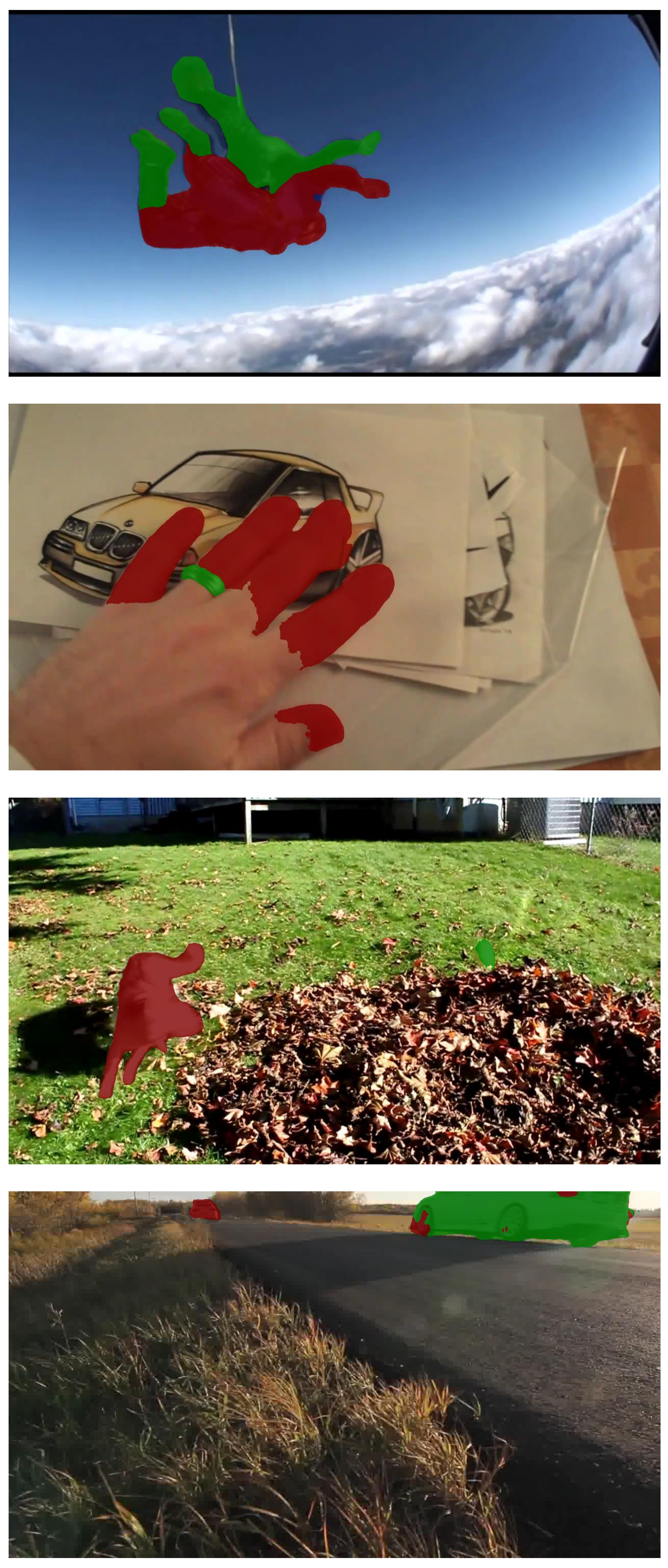}
	\caption{Overlay}
	\end{subfigure} %
	\caption{\textcolor{black}{Visualization of VOS results by different algorithms. Visualization examples are from Youtube-VOS validation set. Different objects are masked in different colors. Overlays of segmentation results by our method and RGB images are shown on the right. Note that ground truth segmentation masks are withheld in Youtube-VOS validation set.}}
	\label{fig:vos-yt}
\end{figure*}

\subsubsection{YouTubeVOS}
VOS methods are evaluated on Youtube-VOS validation set, as shown in Table \ref{ytvalidation}. They are trained on DAVIS and Youtube-VOS training sets. {\color{black}Notably,} {\bf{seg}}, {\bf{syn}}, and $\diamond$ share the same meaning as in Table \ref{dvvalidation}. {\bf{S}} and {\bf{unS}} are seen and unseen object categories, and {\bf{all}} indicates the average of overall $\mathcal{J}$ and $\mathcal{F}$.

Compared with {\color{black}state-of-the-art} methods, our methods achieve the second and third best results. The FAMINet-1F and FAMINet-2F improve the baseline method FRTM-fast$\diamond$\cite{robinson2020learning} (appearance network) by 1.3 percent and 2.2 percent, respectively. Some visualization results are shown in Figure \ref{fig:vos-yt}. Compared {\color{black}with} the DAVIS benchmark, more small objects {\color{black}are present} in YoutubeVOS, which increase the difficulty of VOS. The baseline FRTM-fast \cite{robinson2020learning} fails in segmenting these small objects (e.g., the ball on the third row, and the distant car on the fourth row), while our method segments them accurately. Our method also preserves better object boundaries than a few existing VOS methods (RANet\cite{wang2019ranet} and TTVOS \cite{park2020ttvos}).

\begin{figure}[t] \centering
\setlength{\belowcaptionskip}{-0.2cm}
	\begin{subfigure}[b]{0.49\linewidth}
		\centering
		\includegraphics[width=40mm]{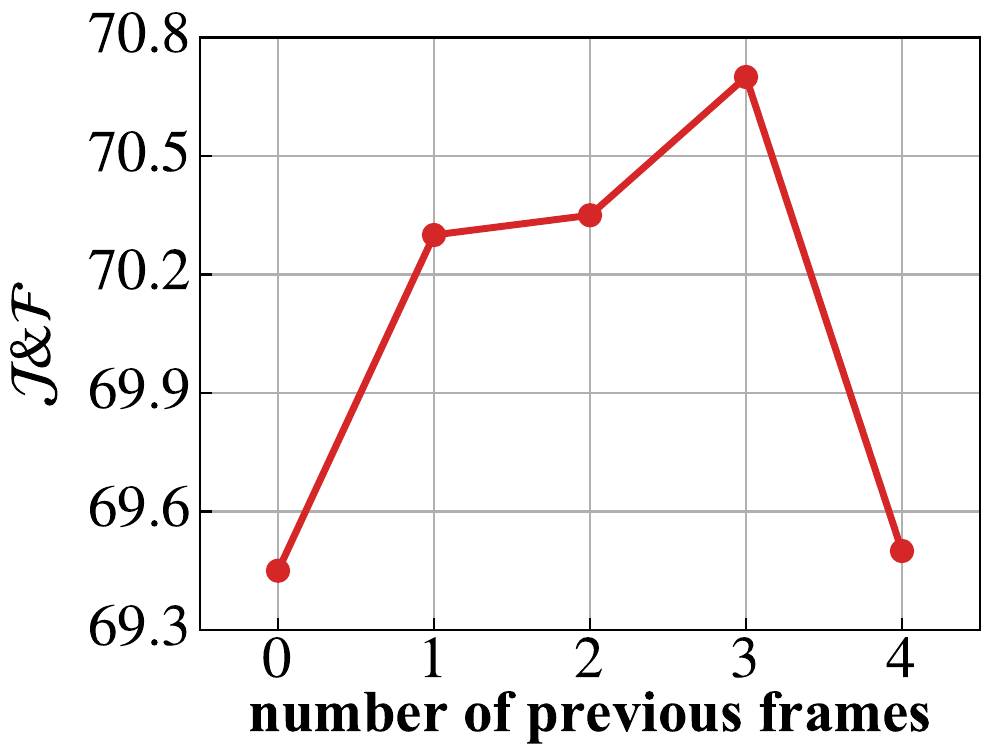}
		\caption{seen}
	\end{subfigure} %
		\begin{subfigure}[b]{0.49\linewidth}
		\centering
		\includegraphics[width=40mm]{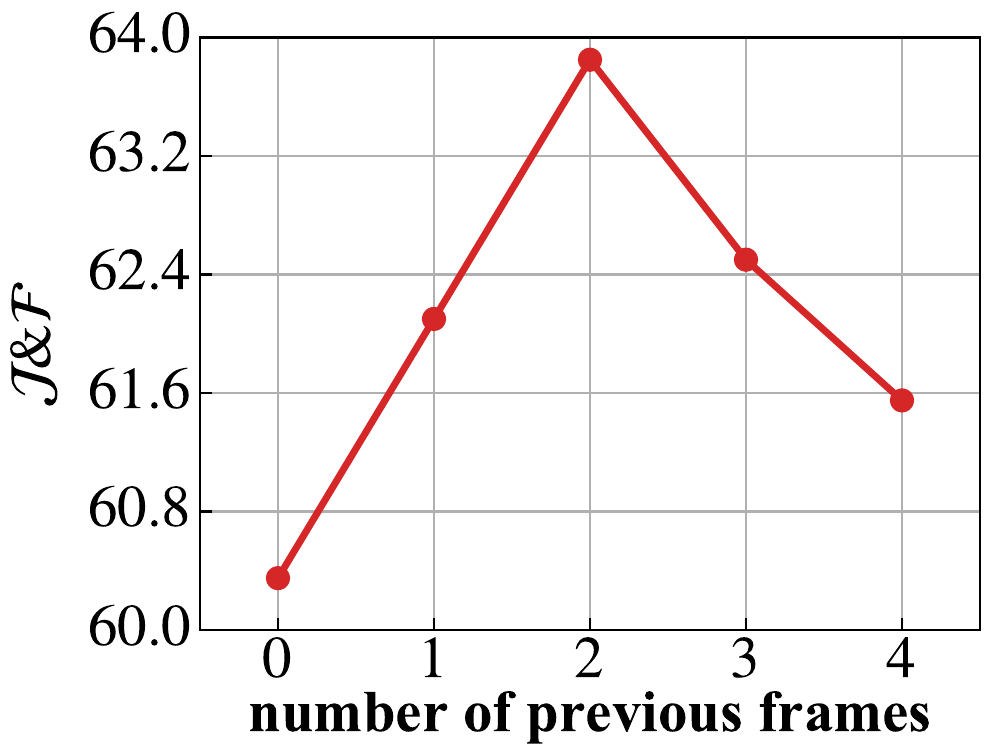}
		\caption{unseen}
	\end{subfigure} %
	
	\caption{Impact of utilizing different numbers of previous frames. Experiments are done on Youtube-VOS validation set, which is split into two sets according to {\color{black}``seen''} and {\color{black}``unseen''} categories. When utilizing 0 previous frames, the FAMINet degrades to the baseline method FRTM-fast.}
	\label{fig:number}
\end{figure}

\begin{figure}[t] \centering
	\begin{subfigure}[b]{0.98\linewidth}
		\centering
		\includegraphics[width=85mm]{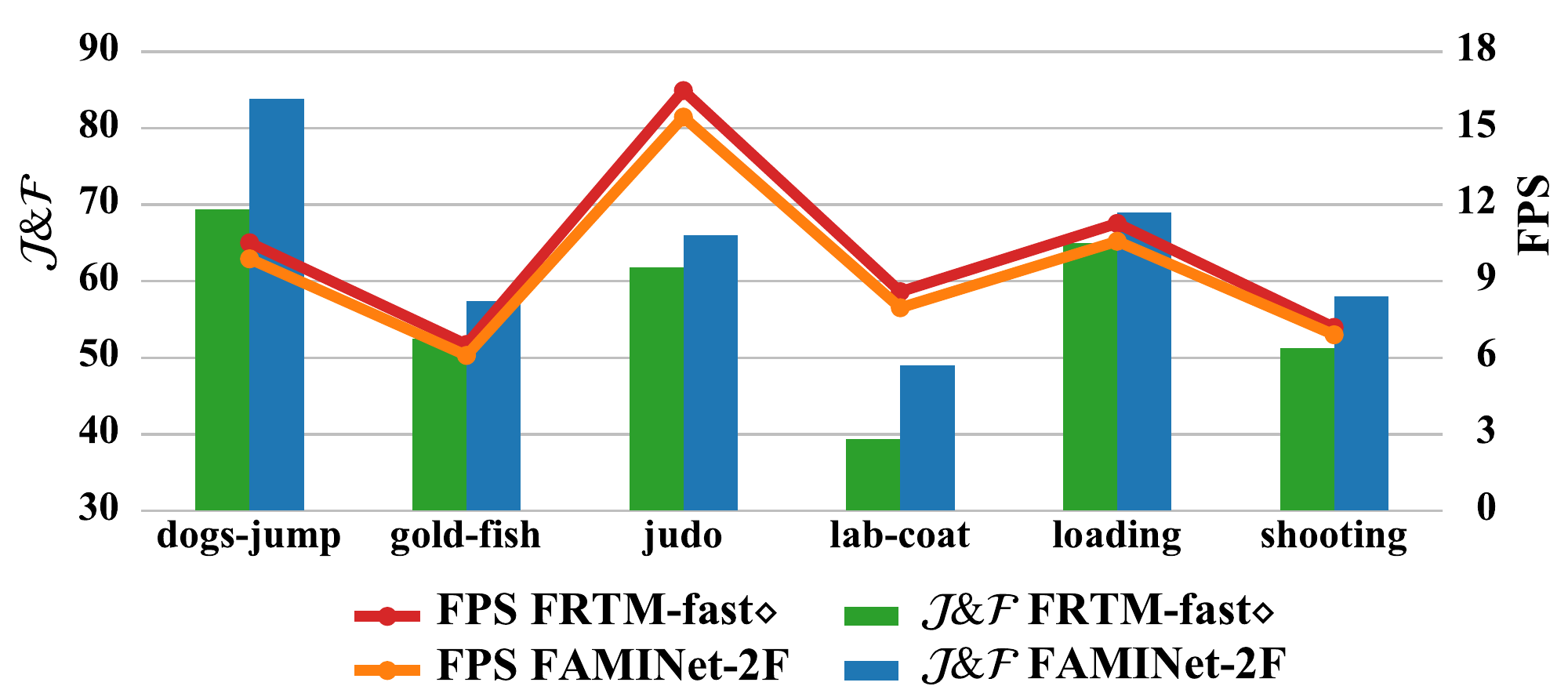}
		\caption{DAVIS 2017 validation set}
	\end{subfigure} %
		\begin{subfigure}[b]{0.98\linewidth}
		\centering
		\includegraphics[width=85mm]{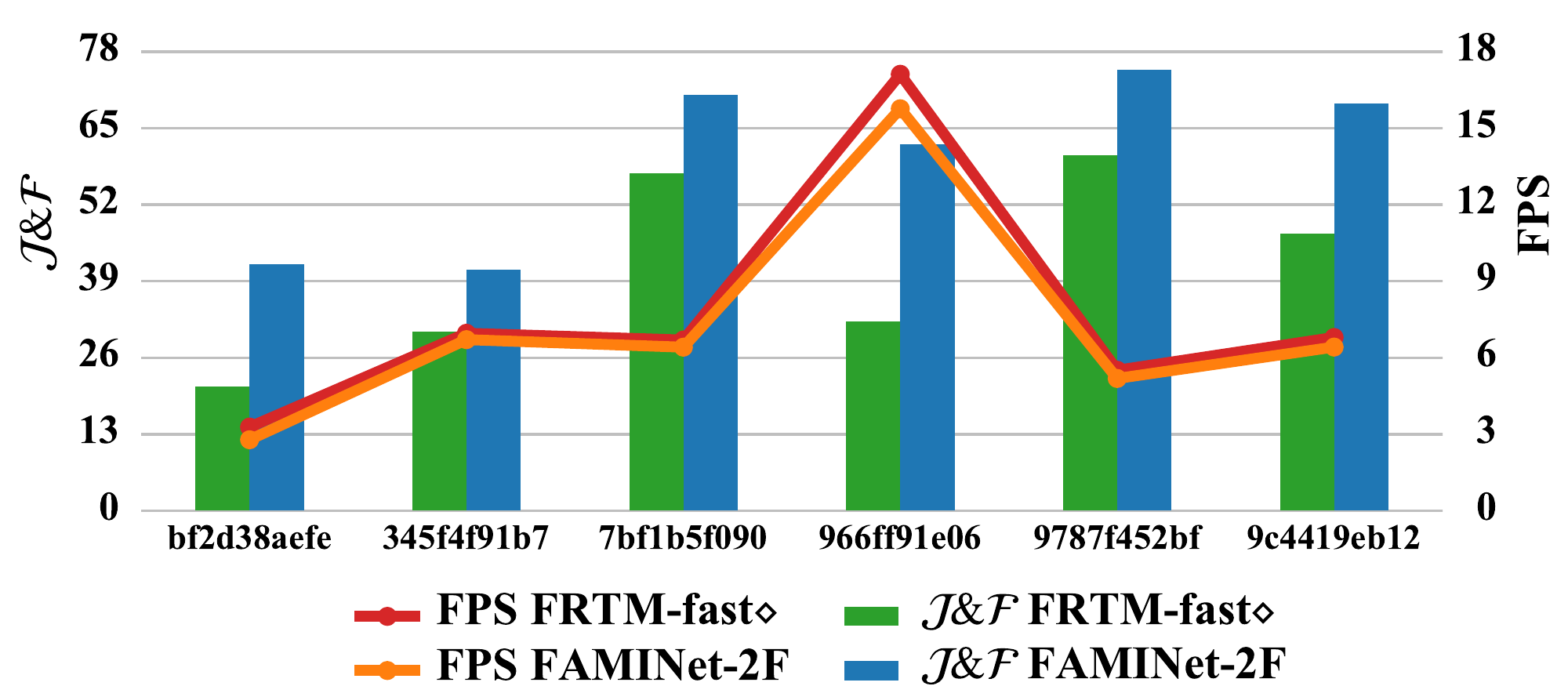}
		\caption{Youtube-VOS validation set}
	\end{subfigure} %
	
	\caption{Accuracy ($\mathcal{J}$\&$\mathcal{F}$) and efficiency (FPS) comparison between the baseline method FRTM-fast and proposed method FAMINet-2F. Compared examples are from DAVIS 2017 and Youtube-VOS validation sets.}
	\label{fig:comparevideo}
\setlength{\belowcaptionskip}{-2cm}
\end{figure}

\begin{figure*}[t]
	\centering
	\includegraphics[width=150mm]{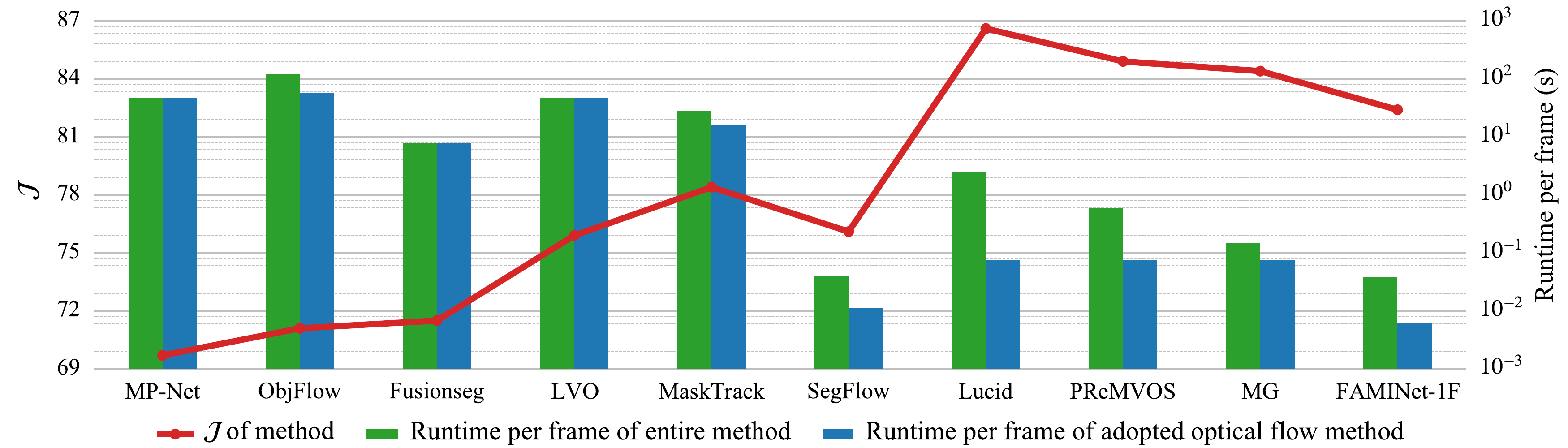}
	\caption{\textcolor{black}{Accuracy ($\mathcal{J}$) and efficiency (Runtime per frame) comparison on DAVIS 2016 validation set. The compared VOS methods are all optical flow-based.}}
	\label{fig:ofvos-compared}
\end{figure*}

\begin{figure}[t] \centering
	\begin{subfigure}[b]{0.98\linewidth}
		\centering
		\includegraphics[width=85mm]{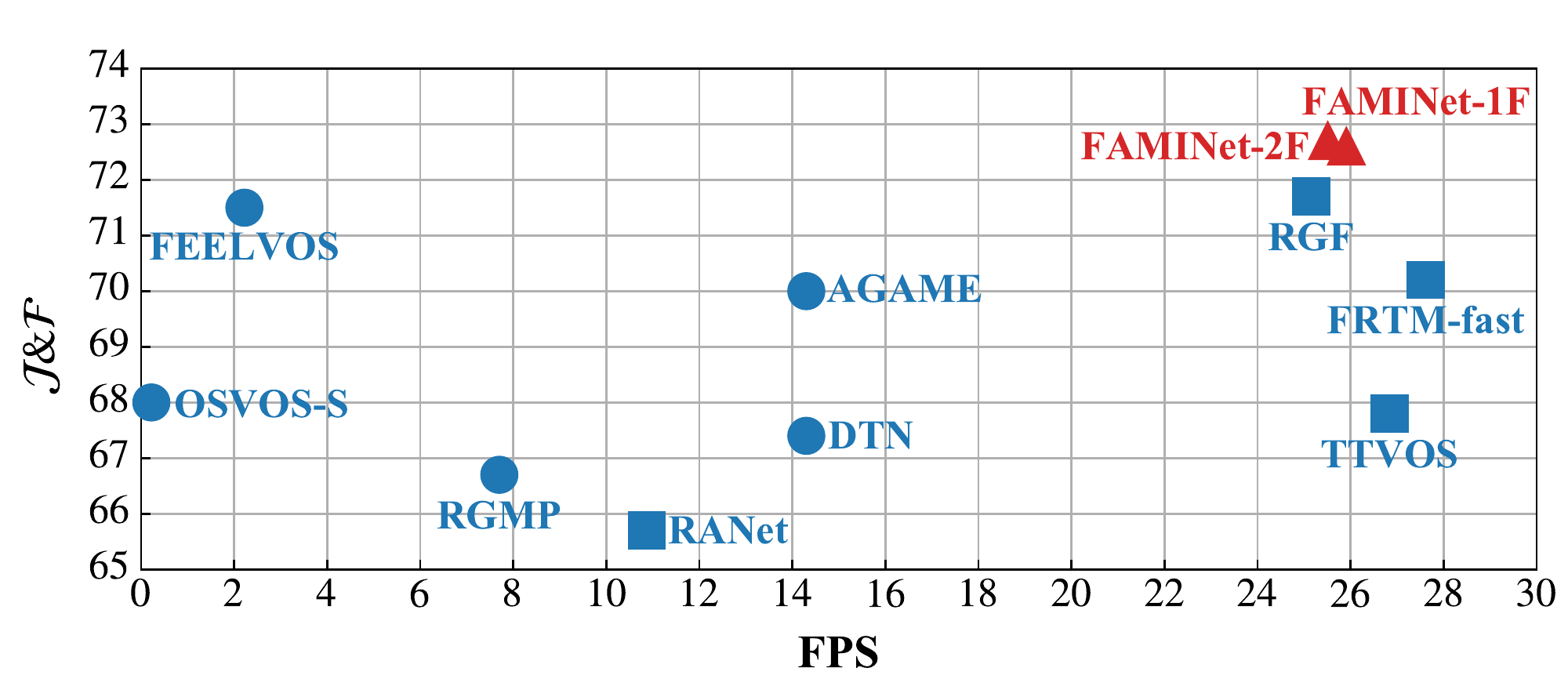}
	\end{subfigure} %
	\caption{$\mathcal{J}$\&$\mathcal{F}$ {\color{black}vs.} FPS. $\mathcal{J}$\&$\mathcal{F}$ is measured on DAVIS 2017 validation set, while the FPS is measured on DAVIS 2016 validation set. $\bigcirc$ indicates that the results of methods are quoted from their original papers. $\Box$ indicates that the results are re-implemented in our environment. $\triangle$ {\color{black}indicates} the results of our proposed methods.}
	\label{fig:fpsjf}
\end{figure}

We further investigate the performances of FAMINet utilizing different numbers of previous frames on the Youtube-VOS benchmark, as shown in Figure \ref{fig:number}. \textcolor{black}{Introducing more than 3 or 2 frames can hurt the segmentation results, which is the same as on DAVIS validation set according to Section \ref{sec:ablation}.} {\color{black}The reason is that capturing} object motions across far apart video frames {\color{black}is difficult}.

\subsection{\textcolor{black}{Analysis of accuracy and efficiency}}
\subsubsection{\textcolor{black}{Compared with the baseline VOS method}}
\textcolor{black}{The FAMINet is compared with the baseline method FRTM-fast\cite{robinson2020learning} on a few video sequences from the DAVIS and YouTube-VOS datasets, as shown in Figure \ref{fig:comparevideo}. Our method improves the accuracy significantly on videos containing multi objects, while it sacrifices little on efficiency.}
\subsubsection{\textcolor{black}{Compared with the optical flow-based VOS methods}}
\textcolor{black}{We measure the running times of several state-of-the-art optical flow-based VOS methods to evaluate their efficiency. The averaged runtime per frame is measured from two aspects: a) the time of running the entire VOS method (including the optical flow method); b) the time of running the optical flow method only. The results are shown in Figure \ref{fig:ofvos-compared}. We can see that traditional optical flow methods take up too many computations, and they result in a very low running speed. Compared with Lucid \cite{khoreva2019lucid}, PReMVOS \cite{luiten2018premvos}, and MG \cite{hu2020motion}, our proposed FAMINet gives up a little accuracy (approximately 2-4 percent on $\mathcal{J}$) but takes up significantly fewer computations (about 10-100 times less on runtime). Our proposed optical flow method is also significantly faster than traditional and deep learning optical flow methods.}
\subsubsection{\textcolor{black}{Compared with the state-of-the-art VOS methods}}
\textcolor{black}{We plot the results of FAMINet and state-of-the-art VOS methods in Figure \ref{fig:fpsjf}. The FAMINet achieves a better trade-off between accuracy and efficiency than existing VOS methods. It also achieves equivalent or even better performance on FPS compared with these optical flow-free methods.}

\section{Conclusion}
\textcolor{black}{In this study, a novel CNN-based method, the FAMINet, is proposed to study the VOS with the help of optical flow. The FAMINet includes a feature extraction network (F) for extracting robust feature representations, an appearance network (A) for initial segmentation, a motion network (M) for estimating the optical flow, and an integration network (I) for refining segmentation. An RSD algorithm is also proposed to optimize the motion network online via a fast speed.} \textcolor{black}{The RSD has a wider range of applications than the original SD, and it leads to a faster convergence speed than the commonly used SGD. The proposed motion and integration networks can be combined with other state-of-the-art VOS methods to improve their performances.} The proposed motion network can be extended to study ghost removal for differently exposed images \cite{zheng2013hybrid}. This problem will be studied in our future research.

\appendices

\ifCLASSOPTIONcaptionsoff
  \newpage
\fi

\balance
\bibliographystyle{ieeetr}
\bibliography{mybib}

\end{document}